\documentclass[runningheads]{llncs}
\usepackage{graphicx}
\usepackage{amsmath,amssymb} % define this before the line numbering.
\usepackage{color}
\usepackage{multirow}
\usepackage{graphicx}
\usepackage{subfigure}
\usepackage{algorithm}
\usepackage{algorithmic}
\usepackage{verbatim}
\usepackage{ulem}
\usepackage{expl3}

\ExplSyntaxOn
\newcommand\latinabbrev[1]{
	\peek_meaning:NTF . {% Same as \@ifnextchar
		#1\@}%
	{ \peek_catcode:NTF a {% Check whether next char has same catcode as \'a, i.e., is a letter
			#1.\@ }%
		{#1.\@}}}
\ExplSyntaxOff
\newcommand\blfootnote[1]{%
	\begingroup
	\renewcommand\thefootnote{}\footnote{#1}%
	\addtocounter{footnote}{-1}%
	\endgroup
}

\begin{document}
	% \linenumbers
	\pagestyle{headings}
	%\mainmatter
	\title{Affinity Derivation and Graph Merge for Instance Segmentation} 
	% Replace with your title
	
	\titlerunning{Affinity Derivation and Graph Merge for Instance Segmentation}
	% Replace with a meaningful short version of your title
	
	\authorrunning{Y. Liu, S. Yang, B. Li, W. Zhou, J. Xu, H. Li and Y. Lu}
	% Replace with shorter version of the author list. If there are more authors than fits a line, please use A. Author et al.
	
	\author{Yiding Liu\inst{1}\and
		Siyu Yang\inst{2}\and
		 Bin Li\inst{3}\and
		  Wengang Zhou\inst{1}\and\\
		 Jizheng Xu\inst{3}\and
		 Houqiang Li\inst{1}\and
		Yan Lu\inst{3}}

	%Please write out author names in full in the paper, i.e. full given and family names. 
	%If any authors have names that can be parsed into FirstName LastName in multiple ways, please include the correct parsing, in a comment to the volume editors:
	%\index{Lastnames, Firstnames}
	%(Do not uncomment it, because you may introduce extra index items if you do that, we will use scripts for introducing index entries...)
	\institute{Department of Electronic Engineering and Information Science\\
		University of Science and Technology of China\\
		\email{liuyd123@mail.ustc.edu.cn \{zhwg,lihq\}@ustc.edu.cn}\and
		Beihang University\\
		\email{yangsiyu@buaa.edu.cn}\and
		Microsoft Research\\
		\email{ \{libin,jzxu,yanlu\}@microsoft.com}
	}
	
	\maketitle
	\begin{abstract}
		We present an instance segmentation scheme based on pixel affinity information, which is the relationship of two pixels belonging to a same instance. In our scheme, we use two neural networks with similar structure. One is to predict pixel level semantic score and the other is designed to derive pixel affinities.
		Regarding pixels as the vertexes and affinities as edges, we then propose a simple yet effective graph merge algorithm to cluster pixels into instances. Experimental results show that our scheme can generate fine grained instance mask.
		With Cityscapes training data, the proposed scheme achieves 27.3 AP on test set.

		\keywords{instance segmentation, pixel affinity, graph merge, proposal-free}
	\end{abstract}

	\section{Introduction}
	\blfootnote{This work was done when Yiding Liu and Siyu Yang took internship at Microsoft Research Asia.}
	
	With the fast development of Convolutional Neural Networks (CNN), recent years have witnessed breakthrough in various computer vision tasks.
	For example, CNN based methods have surpassed humans in image classification \cite{he2015delving}. The rapid progress enables researchers to challenge object detection \cite{erhan2014scalable,huang2017speed,redmon2016you}, semantic segmentation \cite{garcia2017review}, and even instance segmentation \cite{girshick2014rich,hariharan2014simultaneous}.

	Semantic segmentation and instance segmentation try to label every pixel in images. Instance segmentation is more challenging as it also tells which object one pixel belongs to.
	Basically, there are two categories of methods for instance segmentation. The first one is developed from object detection. If one already has results of object detection, i.e. bounding box for each object, one can move one step further to refine the bounding box semantic information to generate instance results. Since the results rely on the proposals from object detection, such category can be regarded as proposal-based methods. The other one is to cluster pixels into instances based on semantic segmentation result. We refer this category as proposal-free methods.
	
	Recent instance segmentation methods advance in both directions. 
	Proposal-based method is usually an extension of object detection frameworks \cite{pinheiro2015learning,liu2016ssd,girshick2015fast}.
	Fully Convolutional Instance-aware Semantic Segmentation (FCIS) \cite{li2017fully} produces position-sensitive feature maps \cite{dai2016instanceFCN} and generates masks through merging features in corresponding areas. Mask RCNN (Mask Region CNN) \cite{he2017mask} extends Faster RCNN \cite{ren2015faster} with another branch to generate masks with different classes.
	Proposal-based methods produce instance-level result in the region of interest (ROI) to make the mask precise. 
	Therefore, the performance highly depends on the region proposal network (RPN) \cite{ren2015faster}, and is usually influenced by the regression accuracy of the bounding box.
	
	Meanwhile, methods without proposal generation have also been developed.
	The basic idea of these methods  \cite{liang2015proposal,jin2016object,fathi2017semantic,de2017semantic} is to learn instance level features for each pixel with CNN, then a clustering method is applied to group the pixels together.
	Sequential group network (SGN) \cite{liu2017sgn} uses CNN to generate features and makes group decisions based on a series of networks.
	
	In this paper, we focus on proposal-free method and exploit semantic information from a new perspective. Similar to other proposal-free methods, we develop our scheme based on semantic segmentation. In addition to using pixel-wise classification results from semantic segmentation, we propose to derive pixel affinity information that tells if two pixels belong to a same object. We design networks to derive those information for neighboring pixels at various scales. Then taking the set of pixels as vertexes and the pixel affinities as the weight of edges, we construct a graph from the output of the network.
	Then we propose a simple graph merge algorithm to group the pixels into instances. More details will be shown in Sec. \ref{subsec:postprocess}.
	By doing so, we can achieve a state-of-the-art result on Cityscapes \textit{test} set with only Cityscapes training data.
	
	Our contributions are multi-fold:
	\begin{itemize}
		\setlength{\parskip}{0pt}
		\setlength{\parsep}{10pt}
		\item[$\bullet$] we introduce a novel proposal-free instance segmentation scheme, where we use both semantic information and pixel affinity information to derive instance segmentation results.
		\item[$\bullet$] we show that even with a simple graph merge algorithm, we can outperform other methods, including proposal-based ones. It clearly shows that proposal-free methods can have comparable or even better performance than proposal-based methods. We hope that our findings can inspire more people to bring instance segmentation to a new level along this direction.
		\item[$\bullet$] we show that semantic segmentation network is reasonably suitable for pixel affinity prediction with only the meaning of the output changed.
		
	\end{itemize}
	\section{Related Work}
	\label{sec:related work}
	Our proposed method is based on CNN for semantic segmentation, and we adapt this to generate pixel affinities.
	Thus, we first review previous works on semantic segmentation, followed by discussing the works on instance segmentation, which is further divided into proposal-based method and proposal-free method.
	
	\textbf{Semantic segmentation: }
	Replacing fully connected layers with convolution layers, Fully Convolutional Networks (FCN) \cite{long2015fully} adapts classification network for semantic segmentation.
	Following this, many works try to improve the network to overcome shortcomings \cite{lin2016efficient,liu2015semantic,zhao2017pyramid}.
	To preserve spatial resolution and enlarge the corresponding respective field, \cite{chen2017deeplab,yu2015multi} introduce dilated/atrous convolution to the network structure.
	To explore multi-scale information, PSPNet \cite{zhao2017pyramid} designs a pyramid pooling structure \cite{grauman2005pyramid,lazebnik2006beyond,liu2015parsenet} and Deeplabv2 \cite{chen2017deeplab} proposes Atrous Spatial Pyramid Pooling (ASPP) to embed contextual information.
	Most recently, Chen \textit{et al.} proposes Deeplabv3+ \cite{chen2018encoder} by introducing encoder-decoder structure \cite{newell2016stacked,lin2017feature,fu2017stacked,islam2017gated} to \cite{chen2017rethinking} and achieves promising performance.
	In this paper, we do not focus on network structure design, and any CNN for semantic segmentation would be feasible for our work.
	
	\textbf{Proposal-based instance segmentation: } 
	These methods exploit region proposal to locate the object and then obtain a corresponding mask exploiting detection models \cite{dai2016r,ren2015faster,liu2016ssd,dai2017deformable}.
	DeepMask \cite{pinheiro2015learning} proposes a network to classify whether the patch contains an object and then generates a mask.
	Multi-task Network Cascades (MNC) \cite{dai2016instance} provides a cascaded framework and decomposes instance segmentation task into three phases including box localization, mask generation and classification. Instance-sensitive FCN \cite{dai2016instanceFCN} extends features to position-sensitive score maps, which contain necessary information for mask proposal, and generates instances combined with objectiveness scores. FCIS \cite{li2017fully} makes the position-sensitive maps further with inside/outside scores to encode information for instance segmentation.
	Mask-RCNN \cite{he2017mask} adds another branch on top of Faster-RCNN \cite{ren2015faster} to predict mask output together with box prediction and classification, achieving excellent performance.
	MaskLab \cite{chen2017masklab} combines Mask-RCNN with position-sensitive scores and shows an improvement on performance.

	\textbf{Proposal-free instance segmentation: }
	These methods often consist of two branches, a segmentation branch and a clustering-purpose branch.
	Pixel-wise mask prediction is obtained by the segmentation output and clustering process aims to group the pixels belong to a certain instance together.
	Liang \textit{et al}. \cite{liang2015proposal} predict the number of instances in an image and instance location for each pixel together with semantic mask. Then they perform a spectral clustering to group pixels.
	Long \textit{et al}. \cite{jin2016object} encode instance relationships to classes and exploit the boundary information when clustering pixels.
	Alireza \textit{et al}. \cite{fathi2017semantic} and Bert \textit{et al}. \cite{de2017semantic} try to learn the embedding vectors to cluster instances.
	SGN \cite{liu2017sgn} tends to propose a sequential framework to group the instances gradually from points to lines and finally to instances, which currently achieves the best performance of proposal-free methods.
	
	\begin{figure}[t]
		
		\setlength{\belowcaptionskip}{-15pt}
		\centering
		\includegraphics[height=4cm]{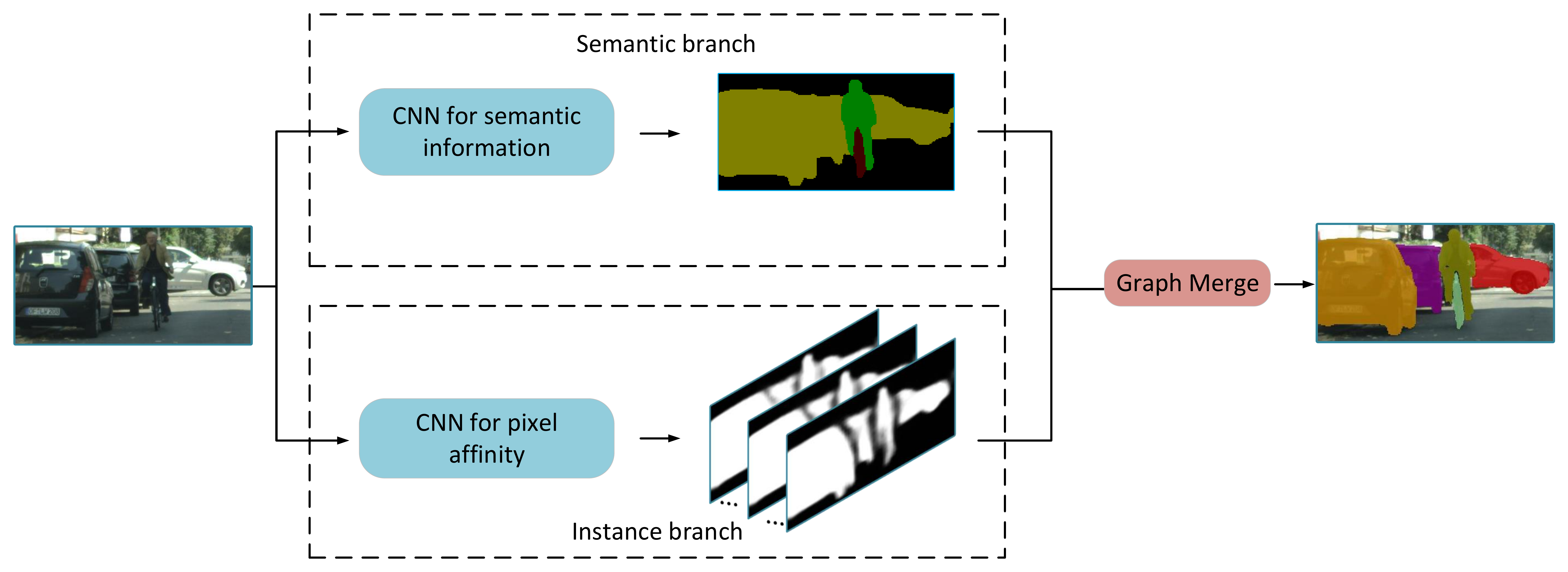}
		\caption{Basic structure for proposed framework.}
		\label{fig:overview}
	\end{figure}
	
	\section{Our Approach}
	\label{sec:method}
	
	\subsection{Overview}
	
	\begin{figure}[t]
		\setlength{\belowcaptionskip}{-15pt}
		\centering 
		\subfigure[]{ 
			%\label{fig:subfig:a} %% label for first subfigure 
			\includegraphics[width=1.1in]{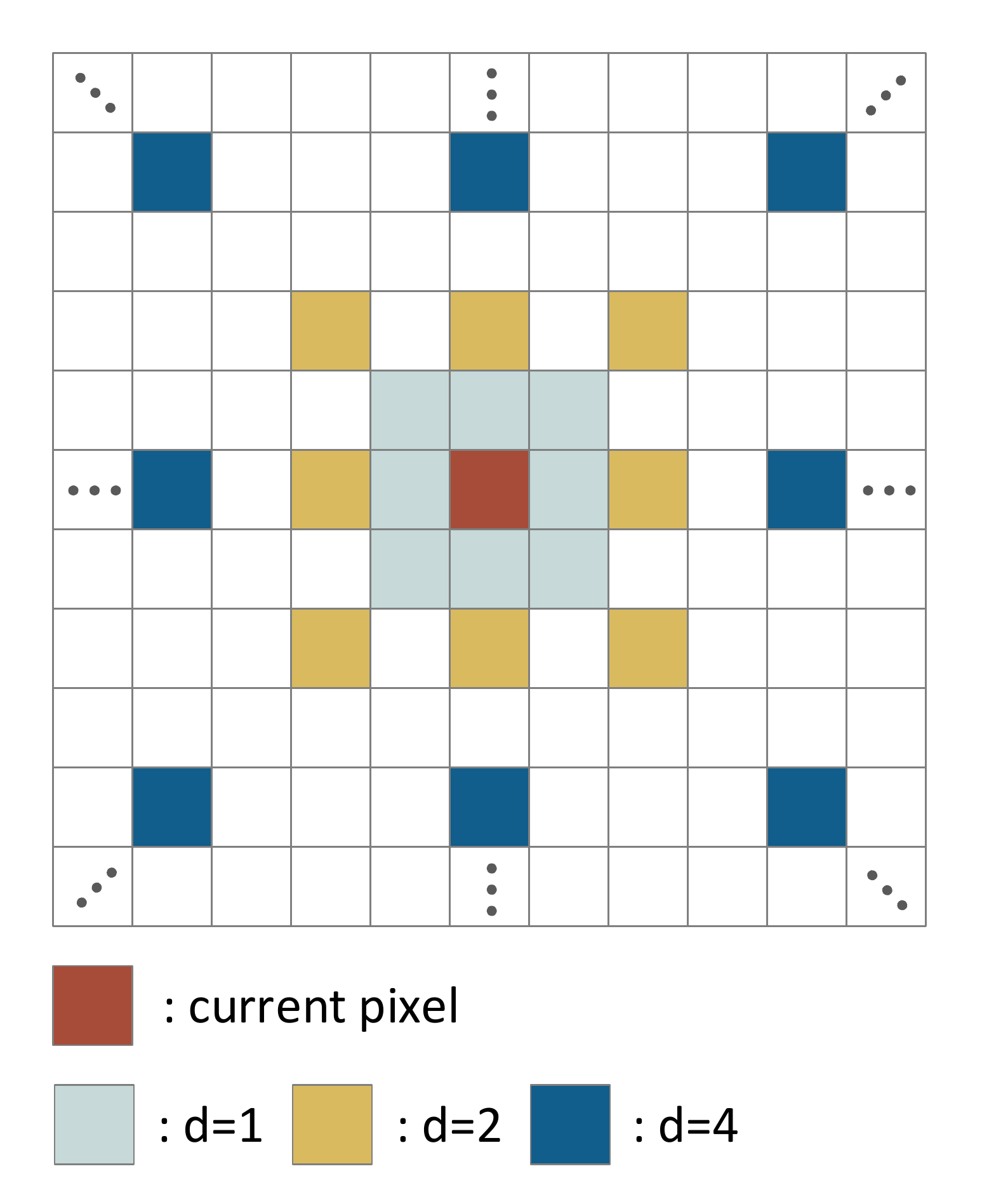}
			\label{fig:neighbor}
		} 
		\subfigure[]{ 
			%\label{fig:subfig:b} %% label for second subfigure 
			\includegraphics[width=1.2in]{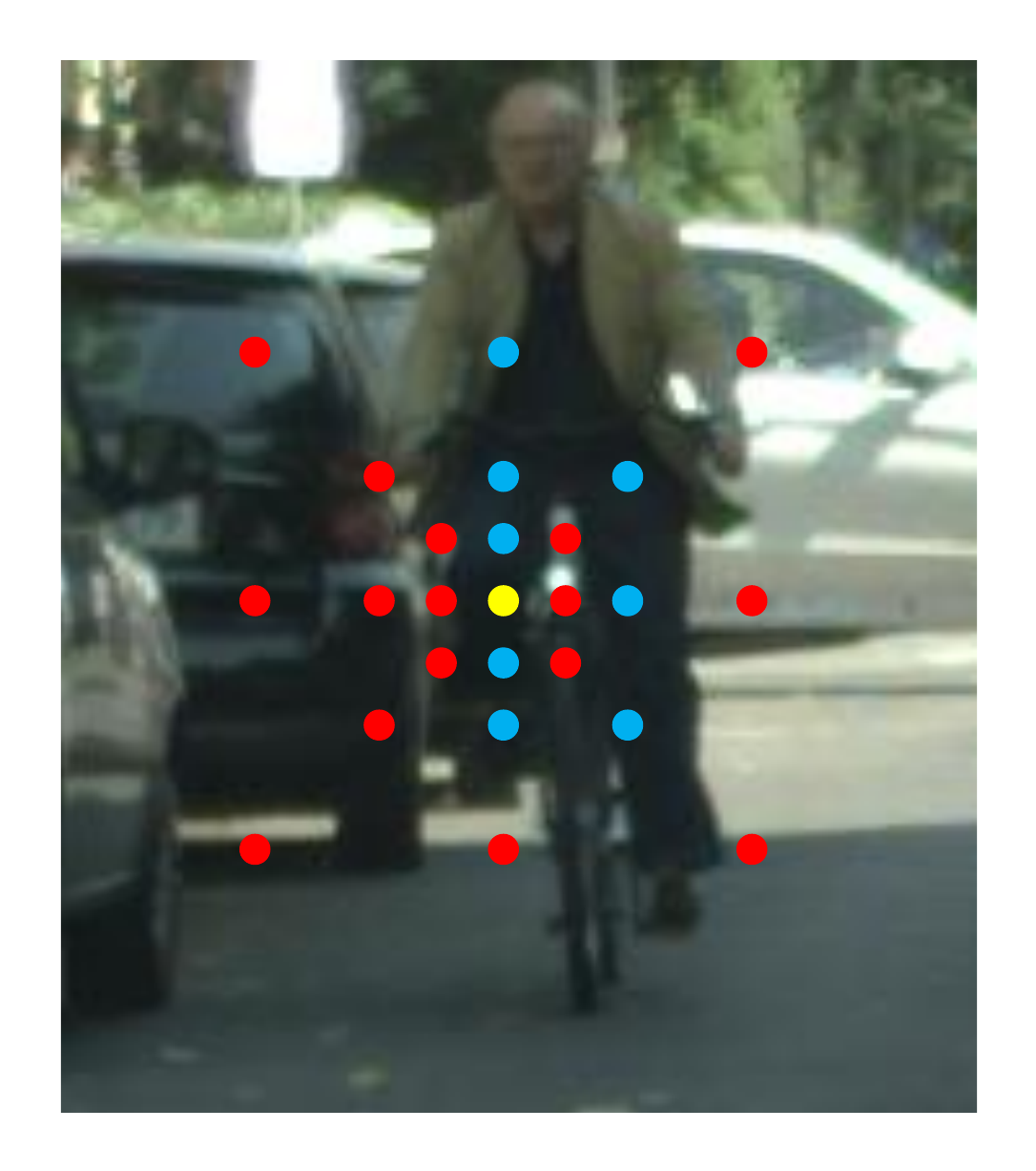}
			\label{fig:sample of neighbor} 
		}
		\subfigure[]{ 
			%\label{fig:subfig:b} %% label for second subfigure 
			\includegraphics[width=1.3in]{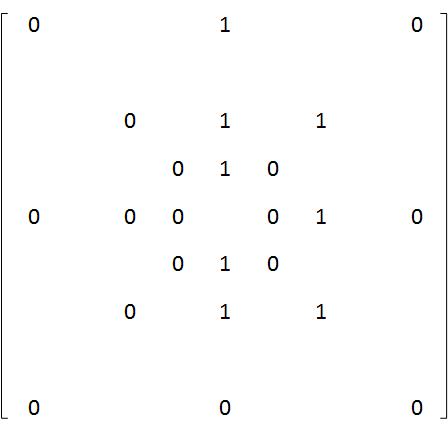}
			\label{fig:neighbor matrix} 
		}
		\label{fig:neighbors and sample}
		\caption{Illustration for pixel affinity. (a) Locations for proposed neighbors. (b) The yellow point indicates the current pixel. Other points are neighboring pixels, in which red ones indicate pixels of different instances and blue ones indicate the same instance (\textit{rider}). The pixel distance is NOT real but only for an illustration. (c) The derived labels and expected network output.}
	\end{figure}
	% \begin{figure}[t]
	%   	\centering
	%    \includegraphics[height=4cm]{inst_v2}
	%    \caption{A illustration for pixel pixel affinity.
	%		    For yellow point pixel, we select two distances, green points represents for the pixels belong to a certain instance and labeled as 1, red ones are relatively labeled as 0 as they belongs to different instances}
	%	    \label{fig:neighbor}
	%    \end{figure}
	The fundamental framework of our approach is shown in Fig. \ref{fig:overview}. 
	We propose to split the task of instance segmentation into two sequential steps. 
	The first step utilizes CNN to obtain class information and pixel affinity of the input image, while the second step applies the graph merge algorithm on those results to generate the pixel-level masks for each instance.
	
	In the first step, we utilize semantic segmentation network to generate the class information for each pixel. 
	Then, we use another network to generate information which is helpful for instance segmentation.
	It is not straightforward to make the network output pixel-level instance label directly, as labels of instance are exchangeable.
	Under this circumstance, we propose to learn whether a pair of neighboring pixels belongs to the same instance.
	It is a binary classification problem that can be handled by the network.
	
	It is impractical to generate affinities between each pixel and all the others in an image. 
	Thus, we carefully select a set of neighboring pixels to generate affinity information. Each channel of the network output represents a probability of whether the neighbor pixel and the current one belong to the same instance, as illustrated in Fig. \ref{fig:neighbor}. 
	As can be seen from the instance branch in Fig. \ref{fig:overview}, the pixel affinities indicate the boundary apparently and show the feasibility to represent the instance information.
	
	In the second step, we consider the whole image as a graph and apply the graph merge algorithm on the network output to generate instance segmentation results. 
	For every instance, the class label is determined by voting among all pixels based on sementic labels.

	\subsection{Semantic Branch}
	\label{subsec: sem branch}
	\begin{figure}[t]
		
		\setlength{\belowcaptionskip}{-15pt}
		\centering
		\includegraphics[height=3.5cm]{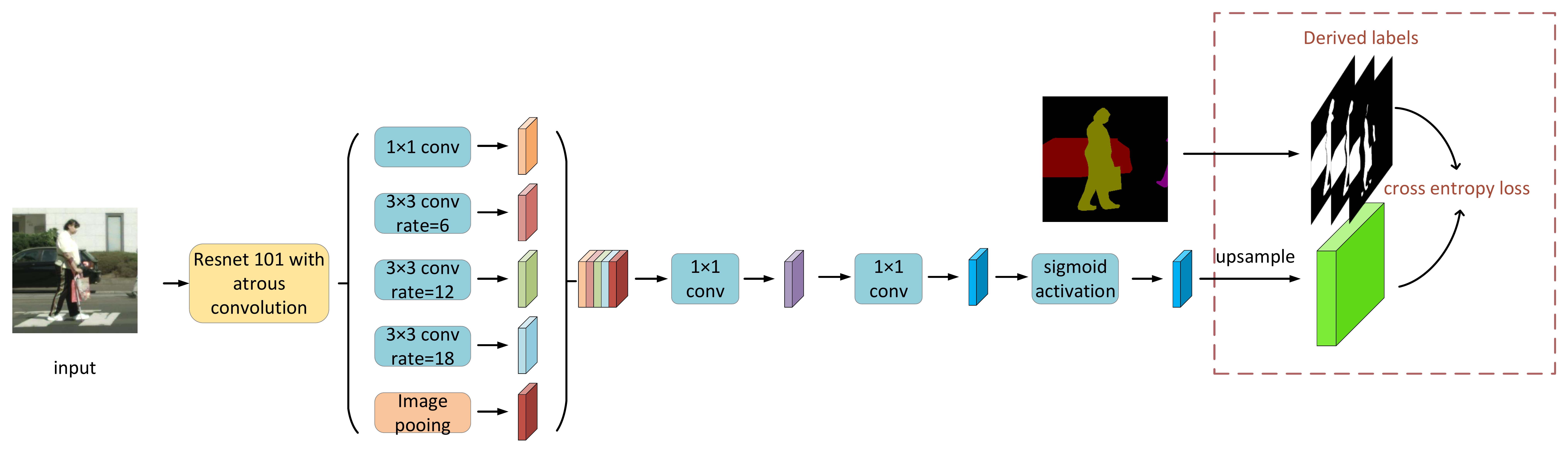}
		\caption{Basic structure for instance branch, we utilize the basic framework from Deeplabv3 \cite{chen2017rethinking} based on Resnet-101 \cite{he2016deep}.}
		\label{fig:deeplabv3}
	\end{figure}
	Deeplabv3 \cite{chen2017rethinking} is one of the state-of-the-art networks in semantic segmentation. Thus, we use it as semantic branch in our proposed framework.
	It should be noted that other semantic segmentation approaches could also be used in our framework. 
	
	\subsection{Instance Branch}
	
	We select several pixel pairs, and the output of instance branch represents whether they belong to the same instance.
	Theoretically, if an instance is composed of only one connected area, we could merge the instance with only two pairs of pixel affinity, {\latinabbrev{i.e}} whether $(p(x,y),p(x-1,y))$ and $(p(x,y),p(x,y-1))$ belong to the same instance, $p(x,y)$ is the pixel at location $(x,y)$ in an image $I$. For the robustness to noise and the ability to handle fragmented instances, we choose the following pixel set as the neighborhood of current pixel $p(x,y)$
	\begin{align}
	N(x,y)=\bigcup_{d\in D}N_d(x,y),
	\end{align}
	where $N_d(x,y)$ is the set of eight-neighbors of $p(x,y)$ with distance $d$, which can be expressed as
	\begin{align}
	N_d(x,y)=\{ p(x+a,y+b), \forall a,b \in \{d,0,-d\}\} \setminus \{p(x,y)\},
	\end{align} 
	and $D$ is the set of distances. In our implementation, $D=\{1,2,4,8,16,32,64\}$, as illustrated in Fig. \ref{fig:neighbor}.
	
	We employ the network in Fig. \ref{fig:deeplabv3} as the instance branch, in which we remove the last softmax activation of semantic segmentation network and minimize the cross entropy loss after sigmoid activation. There are $8\times 7=56$ elements in the set $N(x,y)$, so we assign $56$ channels to the last layer.
	In the training procedure, the corresponding label is assigned as 1 if the pixel pair belongs to a same instance. In the inference procedure, we treat the network outputs as the probability of the pixel pair belonging to the same instance.
	We make a simple illustration of the selected neighbors in Fig. \ref{fig:sample of neighbor}, and the corresponding label is shown in Fig. \ref{fig:neighbor matrix}.
	\subsection{Graph Merge}
	\label{subsec:postprocess}
	\begin{figure}[t]
		\setlength{\belowcaptionskip}{-15pt}
		\centering
		\includegraphics[height=3cm]{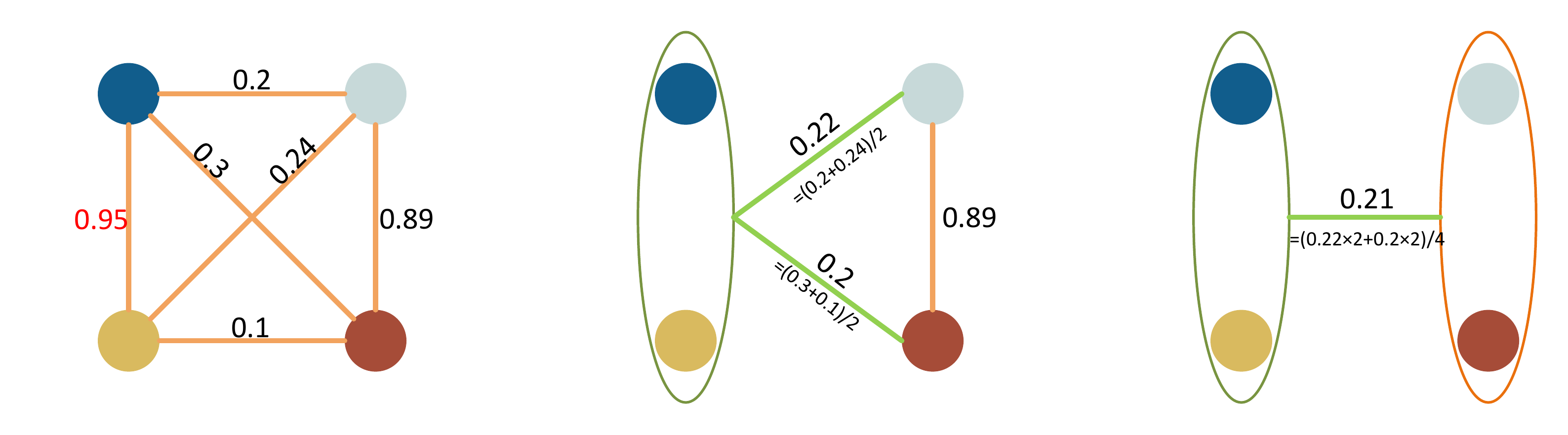}
		\caption{A brief illustration for graph merge algorithm}
		\label{fig:graph merge example}
	\end{figure}
	The graph merge algorithm takes the semantic segmentation and pixel affinity results as input to generate instance segmentation results.
	Let vertex set $V$ be the set of pixels and edge set $E$ be the set of pixel affinities obtained from network. Then, we have a graph $G=(V,E)$. It should be noted that the output of the instance branch is symmetrical. Pair $(p(x,y),p(x_c,y_c))$ obtained at $(x,y)$ and $(p(x_c,y_c),p(x,y))$ at $(x_c,y_c)$ have same physical meaning, both indicating the probability of these two pixels belonging to a certain instance. We average the corresponding probabilities before using them as the initial $E$. Thus, $G$ can be considered as an undirected graph.
	
	Let $e(i,j)$ denote an edge connecting vertex $i$ and $j$. We first find the edge $e(u,v)$ with maximum probability and merge $u,v$ together into a new super-pixel $uv$. It should be noted that we do not distinguish pixel and super-pixel explicitly, and $uv$ is just a symbol indicating it is merged from $u$ and $v$.
	After merging $u,v$, we need to update the graph $G$. For vertex set $V$, two pixels are removed and a new super-pixel is added,
	\begin{align}
	V := V \setminus \{u,v\} \cup \{uv\}.
	\end{align}
	Then, the edge set $E$ needs to be updated. We define $E(u) = \bigcup_{k \in K_u}\{e(u,k)\} $ representing all edges connecting with $u$. $K_u$ is the set of pixels connecting to $u$. $E(u)$ and $E(v)$ should be discarded as $u$ and $v$ have been removed. $K_{uv}=K_u \cup K_v \setminus \{u,v\}$, $E$ is updated as follows,
	\begin{align}
	E := E \setminus E(u) \setminus E(v) \bigcup_{k \in K_{uv}}\{e(uv,k)\}.
	\end{align}
	For $k \in K(u) \cap K(v)$, $e(uv,k)$ is the average of $e(u,k)$ and $e(v,k)$. Otherwise, $e(uv,k)$ inherits from $e(u,k)$ or $e(v,k)$ directly.
	
	After updating $G$, we continue to find a new maximum edge and repeat the procedure iteratively until the maximum probability is smaller than the threshold $r_w$. We summarize the procedure above in Algorithm \ref{alg:simple post processing}.
	We then obtain a set of $V$ and each pixel/super-pixel represents an instance. We recover the super-pixels to sets of pixels and filter the sets with a cardinality threshold $r_c$ which means we only preserve the instance with pixels more than $r_c$.
	We get a set of pixels $X$ as an instance and calculate the confidence of the instance from the initial $E$. We average all the edges $e(i,j)$ for both $i,j \in X$, and this confidence indicates the probability of $X$ being an instance.
	
	{\renewcommand\baselinestretch{0.9}\selectfont
		\begin{algorithm}% enter the algorithm environment
			\caption{Graph Merge Algorithm} % give the algorithm a caption
			\label{alg:simple post processing} % and a label for \ref{} commands later in the document
			\begin{algorithmic}[1] % enter the algorithmic environment
				\REQUIRE Averaged instance branch output $P(u,v)$, thresholds $r_w$
				\ENSURE Merge result $V$, $E$
				\STATE Initialize $V$ with pixels and $E$ with $e(u,v)=P(u,v)$
				\WHILE {Maximum $e(u,v) \in E \geq r_w$}
				\STATE Merge $u,v$ to super-pixel $uv$
				\STATE Update $V$: $V \Leftarrow V \setminus \{u,v\} \cup \{uv\}$
				\STATE $K_{uv}=K_u \cup K_v \setminus \{u,v\}$
				\FOR {$k \in K_{u,v}$}
				\IF {$k \in E(u) \cap E(v)$}
				\STATE $e(uv,k)$ is the average of $e(u,k)$ and $e(v,k)$
				\ELSE
				\STATE $e(uv,k) = k \in K_u ?\ e(u,k):e(v,k)$ 
				\ENDIF
				\ENDFOR
				\STATE Update $E$: $E \Leftarrow E \setminus E(u) \setminus E(v) \bigcup_{k \in K_{uv}} \{e(uv,k)\}$
				\ENDWHILE
			\end{algorithmic}
		\end{algorithm}
		\par}
	We prefer the spatially neighboring pixels to be merged together. Thus, we divide $D=\{1,2,4,8,16,32,64\}$ as three subsets $D_s=\{1,2,4\},\ D_m=\{8,16\}$ and $D_l=\{32,64\}$ with which we do our graph merge sequentially.
	Firstly, we merge pixels with probabilities in $D_s$ with a large threshold $r_{ws}=0.97$, and then all edges with distances in $D_m$ will be added. We continue our graph merge with a lower threshold $r_{wm}=0.7$ and repeat the operation for $D_l$ with $r_{wl}=0.3$.
	
	\section{Implementation Details}
	\label{sec:implementation details}
	The fundamental framework of our approach has been introduced in the previous section. In this section, we elaborate the implementation details.
	
	\subsection{Excluding Background}
	\label{subsec:excluding backgrounds} 
	Background pixels are not necessary to be considered in the graph merge procedure, since they should not be present in any instance. Excluding them decrease the image size as well as accelerate the whole process. We refer the interested sub-regions containing foreground objects as ROI in
	our method. Different from the ROI in proposal-based method, the ROI in our
	method may contain multiple objects.
	In the implementation, we look for connected areas of foreground pixels as ROIs.
	The foreground pixels will be aggregated to super-pixels when generating feasible areas for connecting the separated components belonging to a certain instance. In our implementation, the super-pixel is 32x32, which means if any pixel in a 32x32 region is foreground pixel, we consider the whole 32x32 region as foreground.
	We extend the connected area with a few pixels (16 in our implementation) and find the tightest bounding boxes, which is used as the input of our approach.
	Different from thousands of proposals used in the proposal-based instance segmentation algorithms, the number of ROIs in our approach is usually less than 10.

	\subsection{Pixel Affinity Refinement}
	\label{subsec: semantic weight}
	Besides determining the instance class, the semantic segmentation results can help more with the graph merge algorithm. Intuitively, if two pixels have different semantic labels, they should not belong to a certain instance. Thus, we propose to refine the pixel affinity output from the instance branch in Fig. \ref{fig:overview} by scores from the semantic branch.
	Denote $P(x,y,c)$ as the probability of $p(x,y)$ and $p(x_c,y_c)$ belonging to a certain instance from the instance branch, we refine it by multiplying the semantic similarity of these two pixels.
	
	Let $\mathbf{P}(x,y)=(p_0(x,y),p_1(x,y),\cdots,p_m(x,y))$ denote the probability output of the semantic branch. $m+1$ denotes the number of the classes (including background), $p_i(x,y)$ denotes the probability of the pixel belonging to the $i$-th class and $p_0(x,y)$ is background probability.
	\begin{comment}
	Then, we have $\sum_{i=0}^mp_i(x,y)=1$ as the property of softmax activation. 
	\end{comment}	
	The inner product of the probabilities of two pixels indicates the probability of these two pixels having a certain semantic label. We do not care background pixels, so we discard the background probability and calculate the inner product of $\mathbf{P}(x,y)$ and $\mathbf{P}(x_c,y_c)$ as $\sum_{i=1}^mp_i(x,y)p_i(x_c,y_c)$.
	We then refine the pixel affinity by
	\begin{align}
	P_r(x,y,c) = \sigma(\sum_{i=1}^mp
	_i(x,y)p_i(x_c,y_c))P(x,y,c),
	\end{align}
	
	where
	\begin{align}
	\sigma(x)=2 \times (\frac{1}{1+e^{-\alpha x}}-\frac{1}{2}) \label{equ:mapping function}.
	\end{align}
	This $\sigma()$ function is modified from sigmoid function and we set $\alpha=5$ to weaken the influence of the semantic inner product.
	
	Despite the information we mentioned above, we find that the semantic segmentation model may confuse among classes. Thus, we define the confusion matrix.
	Confusion matrix in semantic segmentation means a matrix where $c_{ij}$ represents the count of pixels belonging to class $i$ classified to class $j$.
	Given this, we can find that the semantic segmentation model sometimes misclassifies a pixel in a subclass, but rarely across sets.
	Thus, we combine classes in each set together as a super-class to further weaken the influence on instance segmentation from the semantic term. Moreover, we set the inner product to 0, when the two pixels are in different super-classes, which helps to refine the instance segmentation result.
	
	\subsection{Resizing ROIs}
	Like what ROI pooling does, we enlarge the shorten edge of the proposed boxes to a fixed size and proportionally enlarge the other edge, with which we use as the input.
	For the Cityscapes dataset, we scale the height of each ROI to 513, if the original height is smaller than it. The reason of scaling it to 513 is that the networks are trained with 513x513 patches. Thus, we would like to use the same value for inference. Moreover, we limit the scaling factor to be less than 4.
	Resizing ROIs is helpful to find more small instances.
	
	\subsection{Forcing Local Merge}
	\begin{figure}[t]
		\setlength{\belowcaptionskip}{-15pt}
		\centering
		\includegraphics[height=2cm]{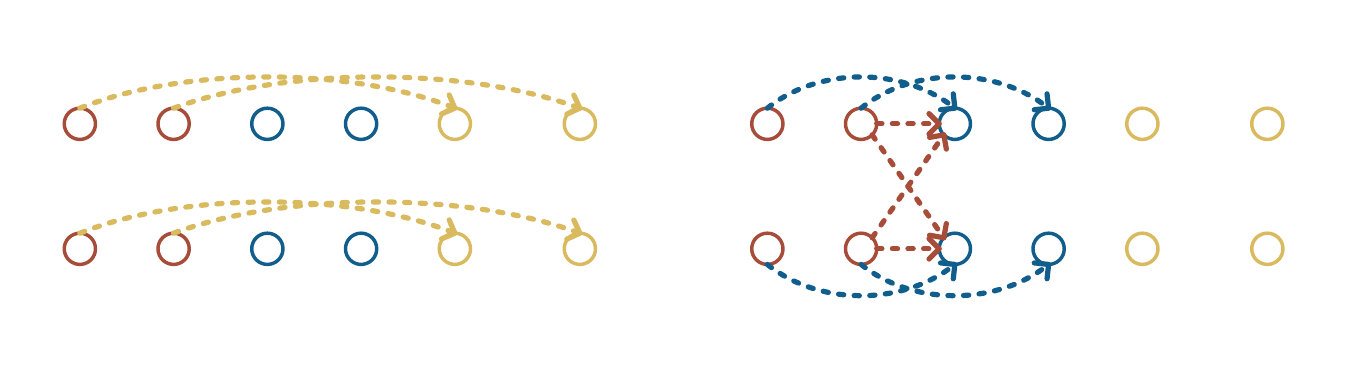}
		\caption{Illustration for forcing local merge. We simulate the merging process with distance $\{1, 2, 4\}$ and window size 2, we only show edges involved in this process. Pixels with identical color are to be merged and we need to update the new weights for edges. Left graph shows that the new probability in distance $\{2\}$ should only be averaged by original weights from distance $\{4\}$ in the same direction. However, the right graph shows the new probability for distance $\{1\}$ should be an average for edges from both distance $\{1\}$ and $\{2\}$.}
		\label{fig:graph merge}
	\end{figure}
	We force the neighboring $m \times m$ pixels to be merged before the graph merge. During the process, we recalculated the pixel affinities according to our graph algorithm in Sec. \ref{subsec:postprocess}. Fig. \ref{fig:graph merge} shows a simple example.
	% add description of the merge process, using 2x2 as example should be OK.
	Force merging neighboring pixels can not only filter out the noises of the network output by averaging, but also decrease the input size of the graph merge algorithm to save processing time. We will provide results on different merge window size in Sec \ref{subsec: results}.

	\subsection{Semantic Class Partition}
	
	To get more exquisite ROIs, we refer to the semantic super-classes in Sec. \ref{subsec: semantic weight} and apply it on the procedure of generating connected areas.
	We sum the probabilities in each super-class and classify the pixels to super-classes. To find foreground region of a super-class, we only consider the pixels classified to this super-class as foreground and all the others as background.
	% The feasible areas will inevitably contains parts of instances from other classes due to the boundary-extending operation when generated. We only preserve the instances which has classes in the specific super-class and discard the other generated instances to avoid repetition and incompleteness. 
	Detailed experimental results will be provided in Sec. \ref{subsec: results}
	
	\section{Experimental Evaluation}
	\label{sec: experimental evaluation}
	
	We evaluate our method on the Cityscapes dataset \cite{Cordts2016Cityscapes}, which consists of $5,000$ images representing complex urban street scenes with the resolution of $2048 \times 1024$.
	Images in the dataset are split into training, validation, and test set of $2,975$, $500$, and $1,525$ images, respectively.
	We use average precision (AP) as our metric to evaluate the results, which is calculated by the mean of IOU threshold from 0.5 to 0.95 with the step of 0.05.
	
	As most of the images in Cityscapes dataset are background on top or bottom, we discard the parts with no semantic labeled pixels on top or bottom for 90\% of training images randomly, in order to make our data more effective.
	To improve the performance of semantic segmentation, we utilize coarse labeled training data by selecting patches containing \textit{trunk, train} and \textit{bus} as additional training data to train the semantic branch.
	We crop 1554 patches from coarse labeled data. To augment data with different scale objects, we also crop several upsampled areas in the fine labeled data. As a result, the final patched fine labeled training data includes 14178 patches, including 2975 original training images with 90\% of them dropped top and bottom background pixels.
	The networks are trained with Tensorflow \cite{abadi2016tensorflow} and the graph merge algorithm is implemented in C++.
	
	\subsection{Training Strategy}
	For the basic setting, the network output strides for both semantic and instance branch are set to 16, and they are trained with input images of size $513 \times 513$.
	
	For the semantic branch, the network structure is defined as introduced in Sec. \ref{subsec: sem branch}, whose weight is initialized with ImageNet \cite{russakovsky2015imagenet} pretrained ResNet-101 model.
	During training, we use 4 Nvidia P40 GPUs with SGD \cite{lecun1989backpropagation} in the following steps.
	(1) We use 19-class semantic labeled data in Cityscapes dataset fine and coarse data together, with initial learning rate of 0.02 and batch size of 16 per GPU. Model is trained using 100k iterations and learning rate is multiplied by 0.7 every 15k iterations.
	(2) As the instance segmentation only focuses on 8 foreground objects, we then finetune the network with 9 classes labeled data (8 foreground objects and 1 background). Training data for this model contains a mix of 2 times fine labeled patched data and coarse labeled patches.
	We keep the other training setting unchanged.
	(3) We finetune the model with 3 times of original fine labeled data together with coarse labeled patches, with other training setting unchanged.
	
	For the instance branch, we still initialize the network parameter with ImageNet pretrained models. We train this model with patched fine labeled training data for 120k iterations, with other settings identical to the step (1) for semantic model training.
	
	\subsection{Main Results}
	As shown in Table \ref{table:result on test}, our method notably improves the performance and achieves 27.3 AP on Cityscapes \textit{test} set, which outperforms Mask RCNN trained with only Cityscapes \textit{train} data by 1.1 points (4.2\% relatively).
	
\begin{figure}[t]
		\setlength{\belowcaptionskip}{-10pt}
		
		\centering
		\begin{minipage}[t]{0.23\textwidth}
			\centering
			\includegraphics[width=2.8cm]{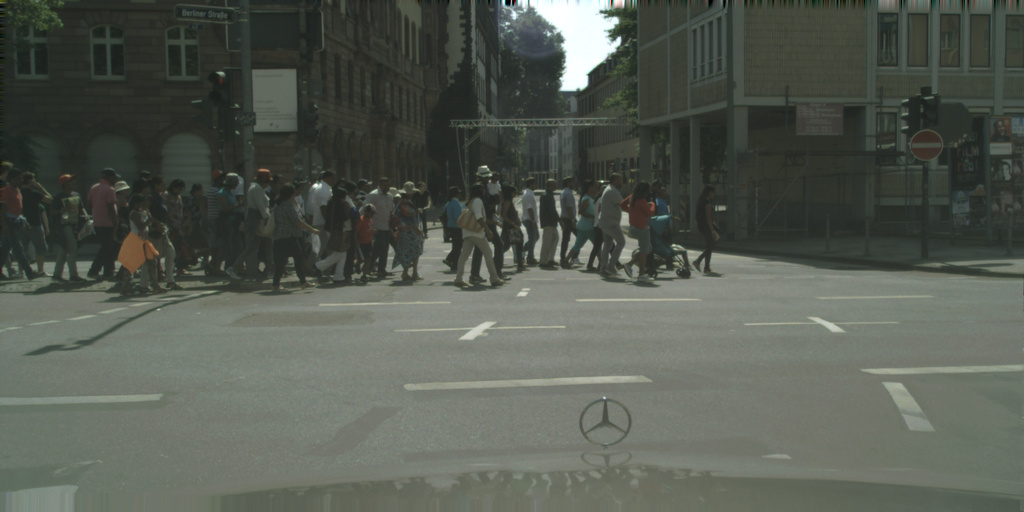}
		\end{minipage}
		\begin{minipage}[t]{0.23\textwidth}
			\centering
			\includegraphics[width=2.8cm]{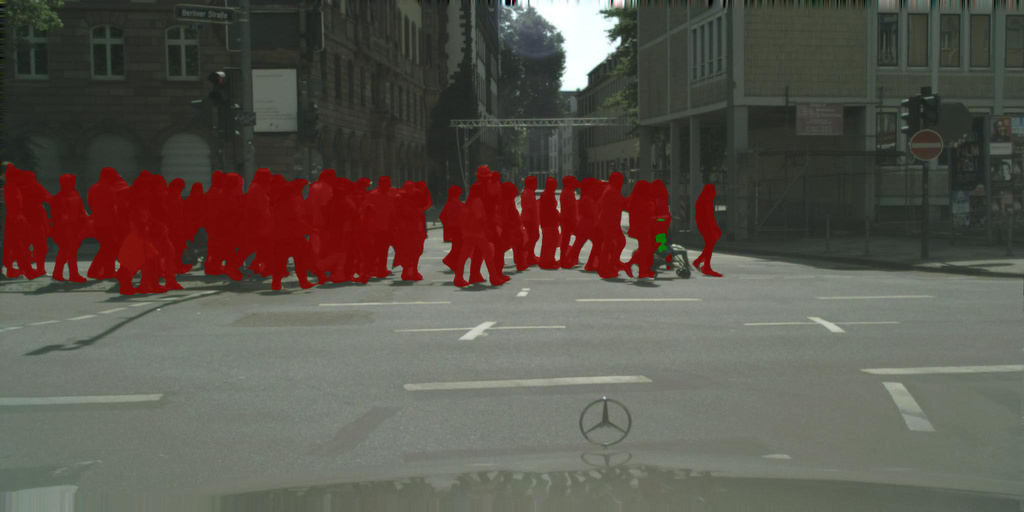}
		\end{minipage}
		\begin{minipage}[t]{0.23\textwidth}
			\centering
			\includegraphics[width=2.8cm]{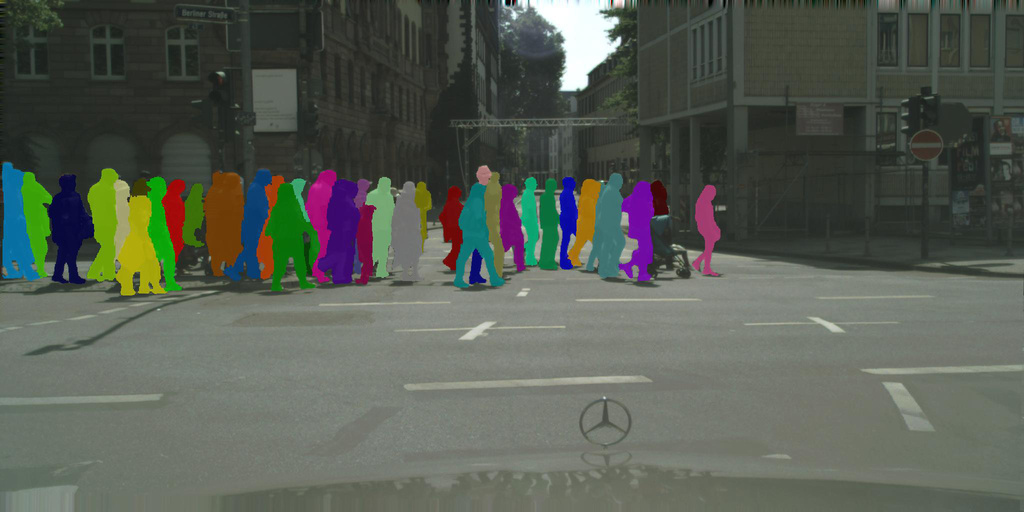}
		\end{minipage}
		\begin{minipage}[t]{0.23\textwidth}
			\centering
			\includegraphics[width=2.8cm]{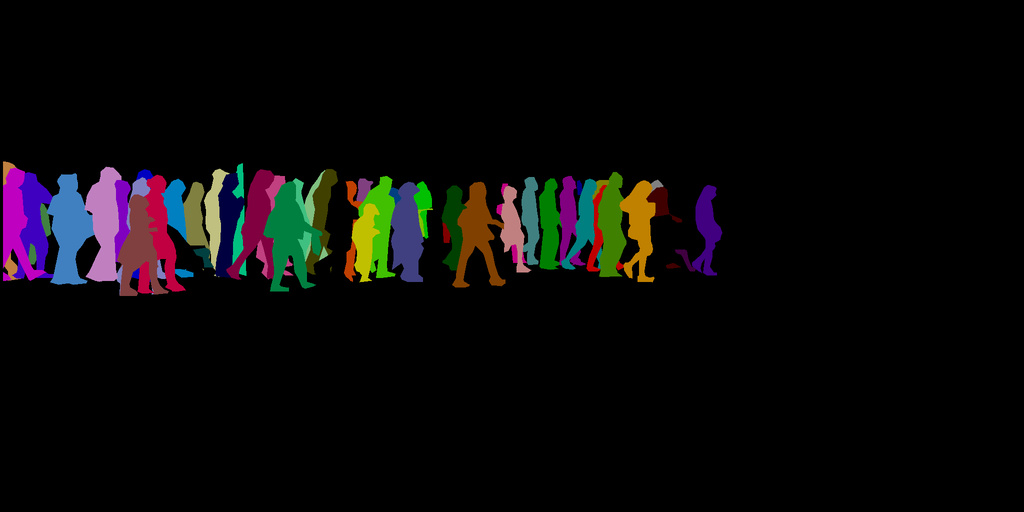}
		\end{minipage}
		\begin{minipage}[t]{0.23\textwidth}
			\centering
			\includegraphics[width=2.8cm]{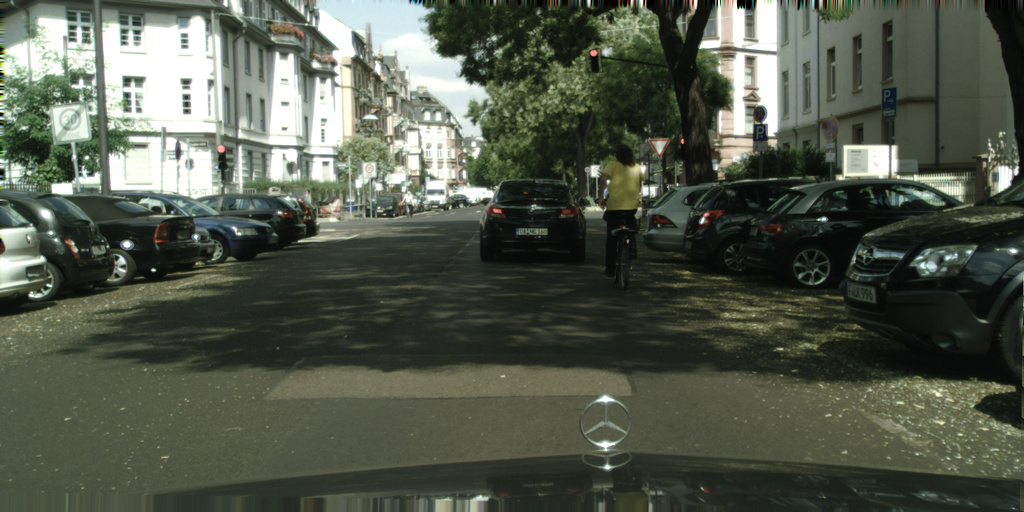}
		\end{minipage}
		\begin{minipage}[t]{0.23\textwidth}
			\centering
			\includegraphics[width=2.8cm]{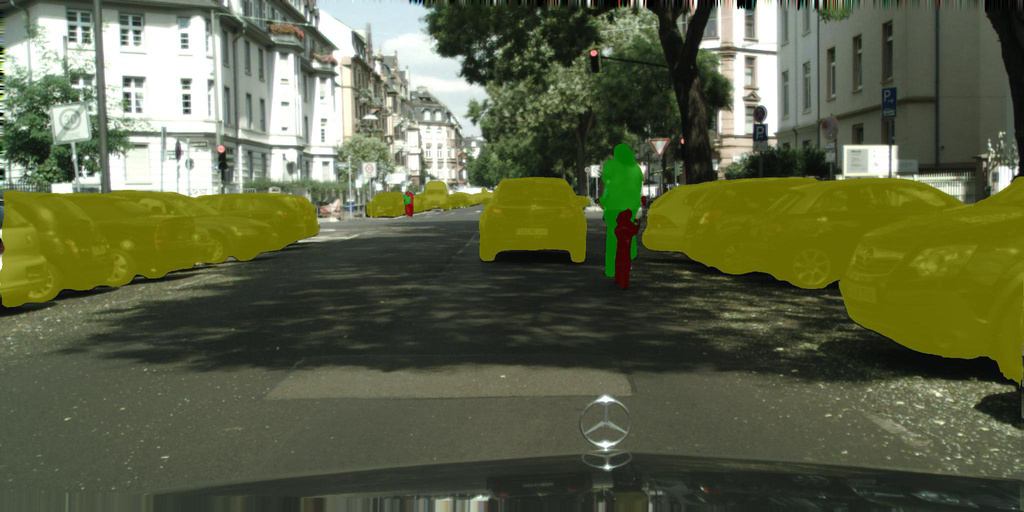}
		\end{minipage}
		\begin{minipage}[t]{0.23\textwidth}
			\centering
			\includegraphics[width=2.8cm]{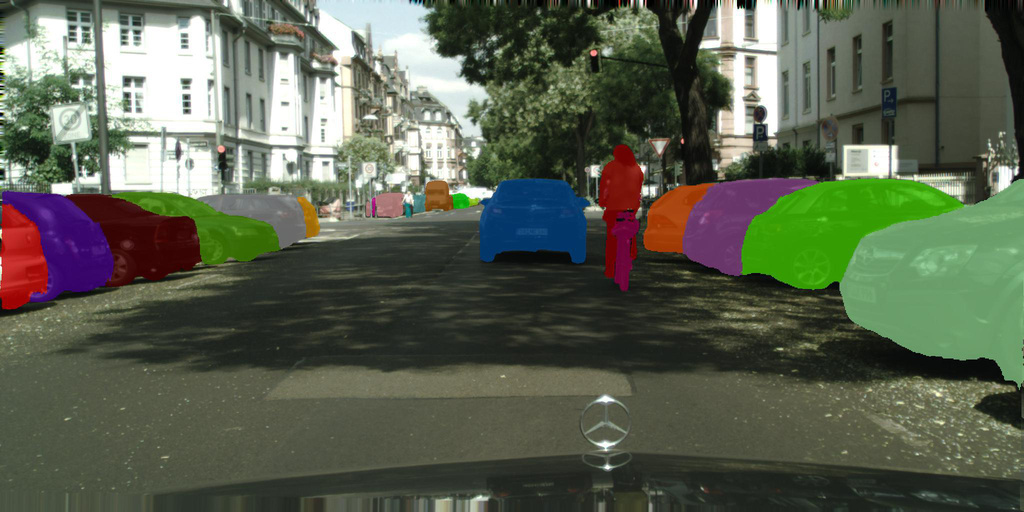}
		\end{minipage}
		\begin{minipage}[t]{0.23\textwidth}
			\centering
			\includegraphics[width=2.8cm]{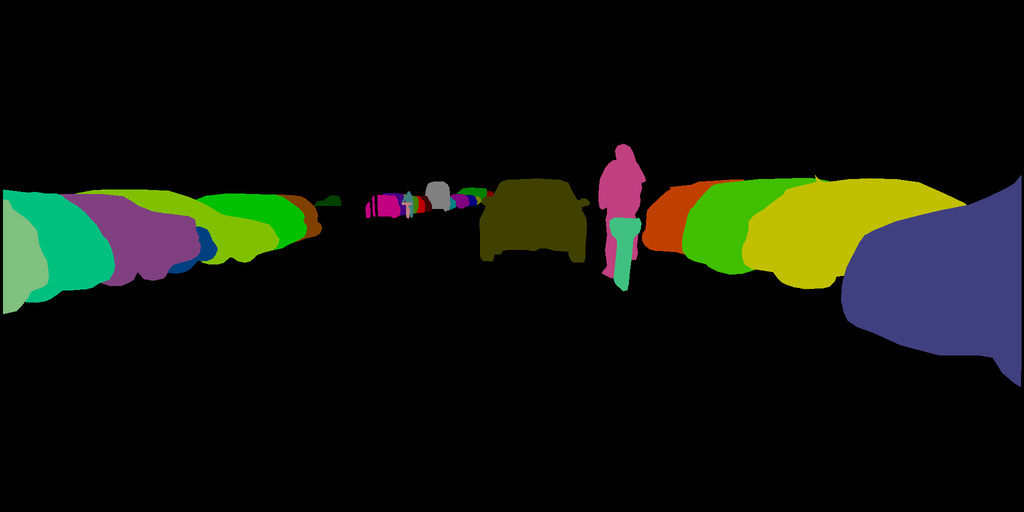}
		\end{minipage}
		\begin{minipage}[t]{0.23\textwidth}
			\centering
			\includegraphics[width=2.8cm]{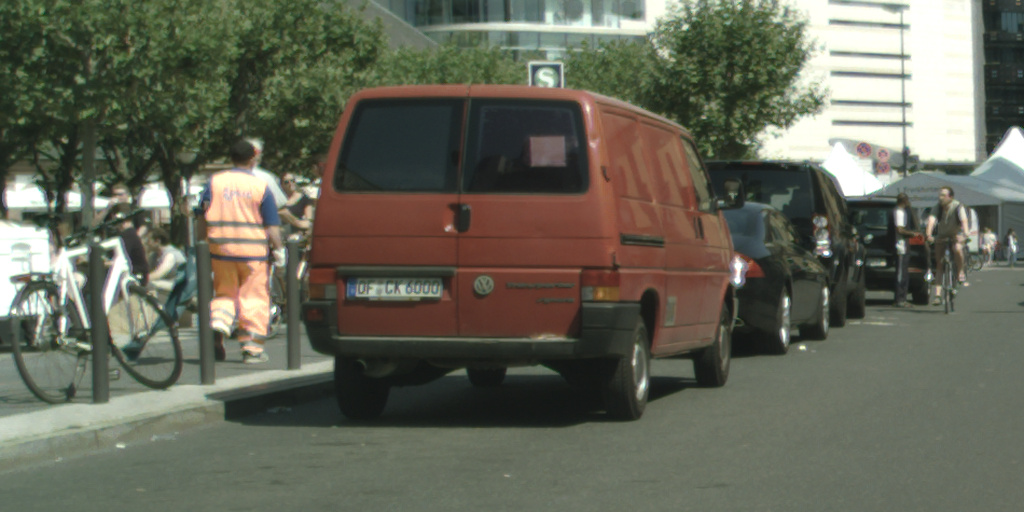}
		\end{minipage}
		\begin{minipage}[t]{0.23\textwidth}
			\centering
			\includegraphics[width=2.8cm]{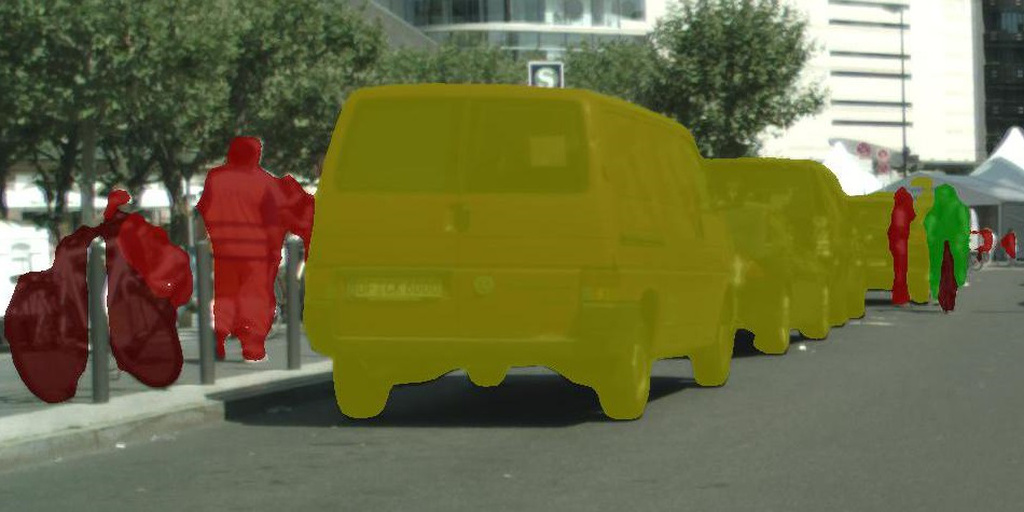}
		\end{minipage}
		\begin{minipage}[t]{0.23\textwidth}
			\centering
			\includegraphics[width=2.8cm]{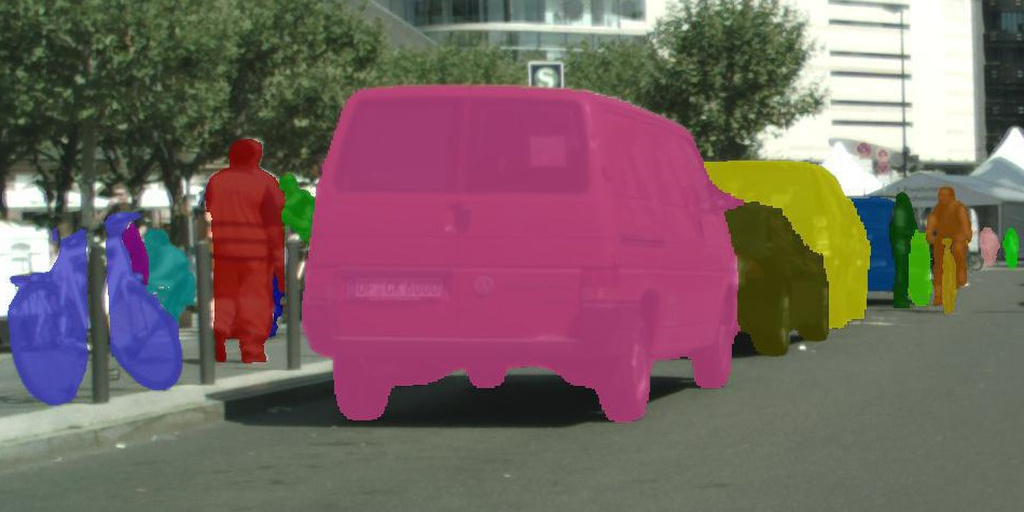}
		\end{minipage}
		\begin{minipage}[t]{0.23\textwidth}
			\centering
			\includegraphics[width=2.8cm]{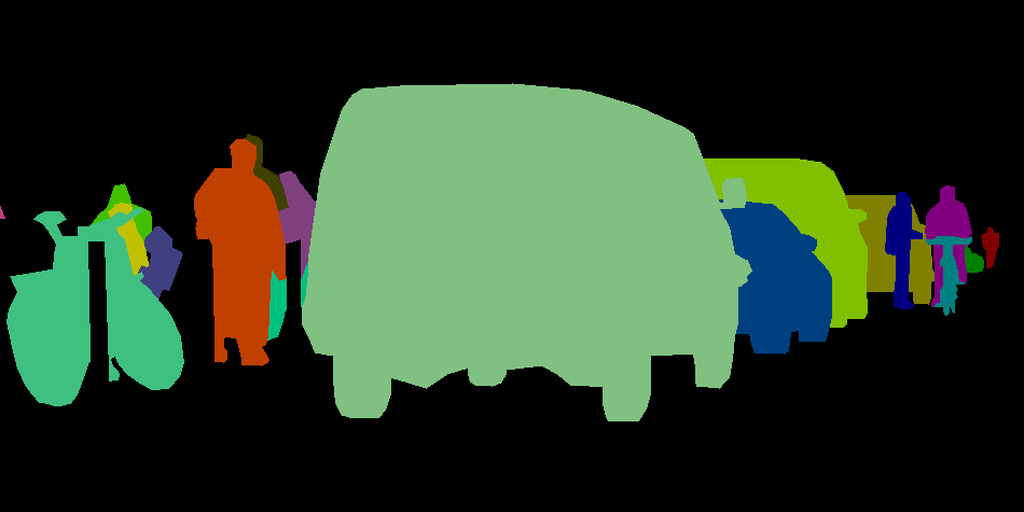}
		\end{minipage}
		\begin{minipage}[t]{0.23\textwidth}
			\centering
			\includegraphics[width=2.8cm]{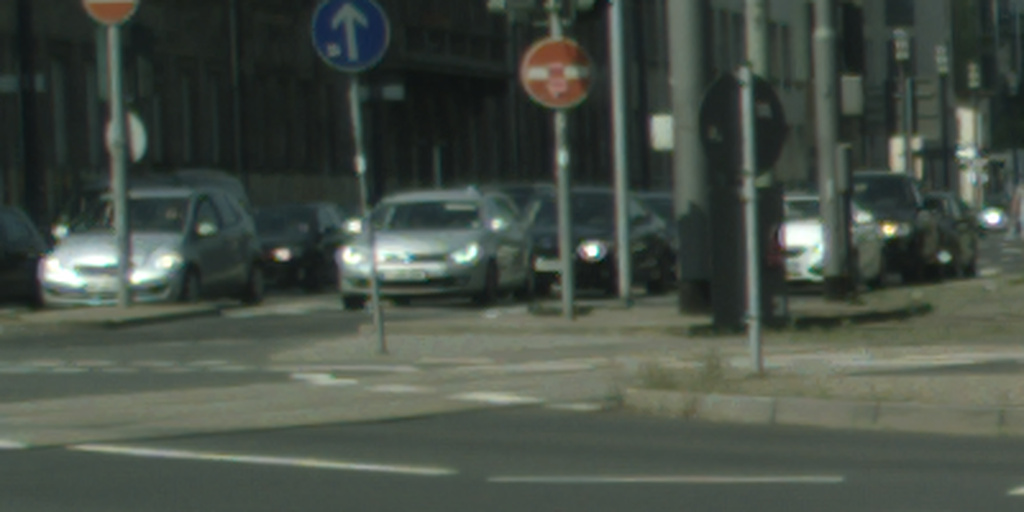}
		\end{minipage}
		\begin{minipage}[t]{0.23\textwidth}
			\centering
			\includegraphics[width=2.8cm]{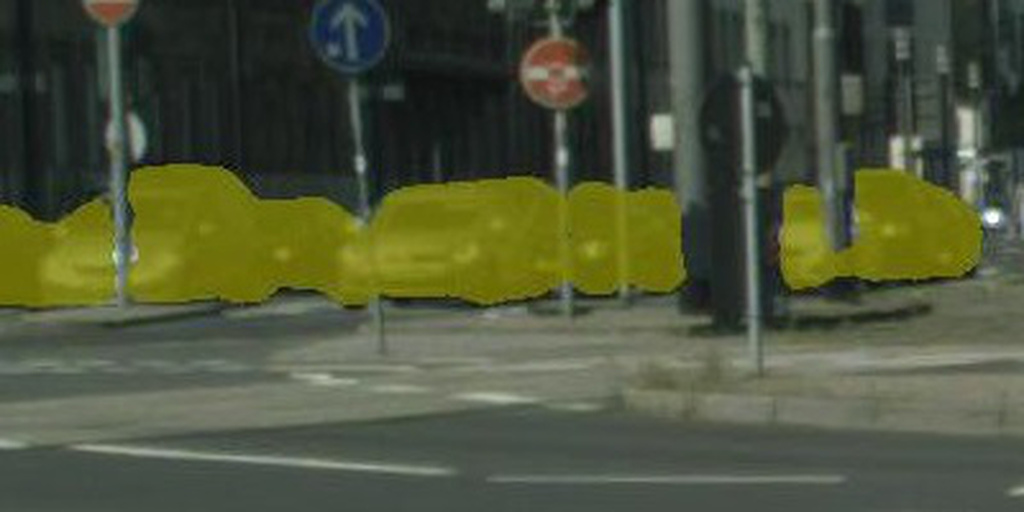}
		\end{minipage}
		\begin{minipage}[t]{0.23\textwidth}
			\centering
			\includegraphics[width=2.8cm]{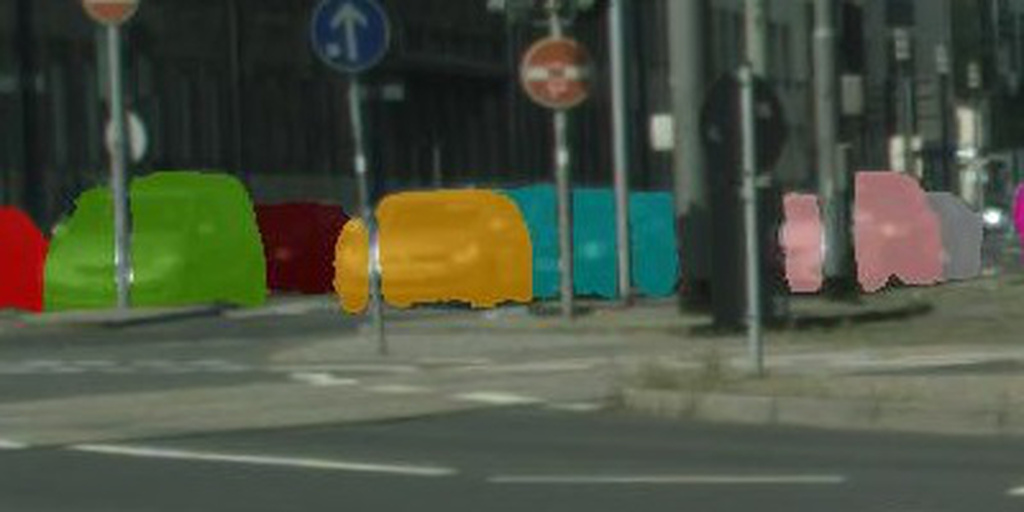}
		\end{minipage}
		\begin{minipage}[t]{0.23\textwidth}
			\centering
			\includegraphics[width=2.8cm]{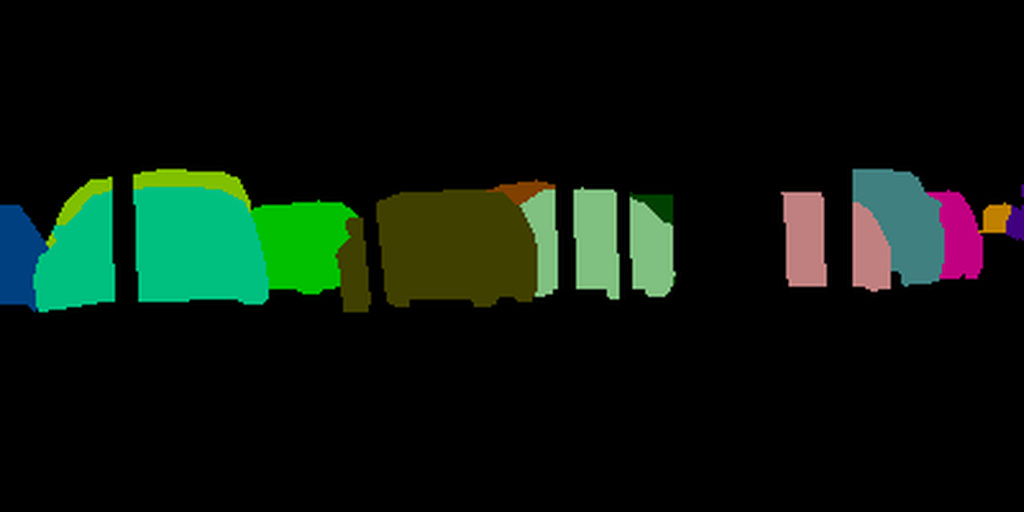}
		\end{minipage}
		
		\caption{Results on Cityscapes \textit{val} dataset, original image, semantic result, instance result and ground truth from left to right. Results in last two rows are cropped from original ones for better visualization.}
		
		\label{fig:results}
	\end{figure}

	We show qualitive results for our algorithm in Fig.
	\ref{fig:results}.
	As shown in the figure, we produce high quality results on both semantic and instance masks, where we get precise boundaries. As shown in the last row of result, we can handle the problem of fragmented instances and merge the separated parts together.
	
	Our method outperforms Mask RCNN on AP but gets a relatively lower performance on AP 50\%. We would interpret it as we could get higher score when the IOU threshold is larger. It means that Mask RCNN could find more instances with relatively less accurate masks (higher AP 50\%), but our method could obtain more accurate boundaries. The bounding box of proposal-based method may lead to a rough mask, which will be judged as correct with small IOU.
	
\begin{figure}[t]
		\setlength{\belowcaptionskip}{-10pt}
		
		\centering
		\begin{minipage}[t]{0.23\textwidth}
			\centering
			\includegraphics[width=2.8cm]{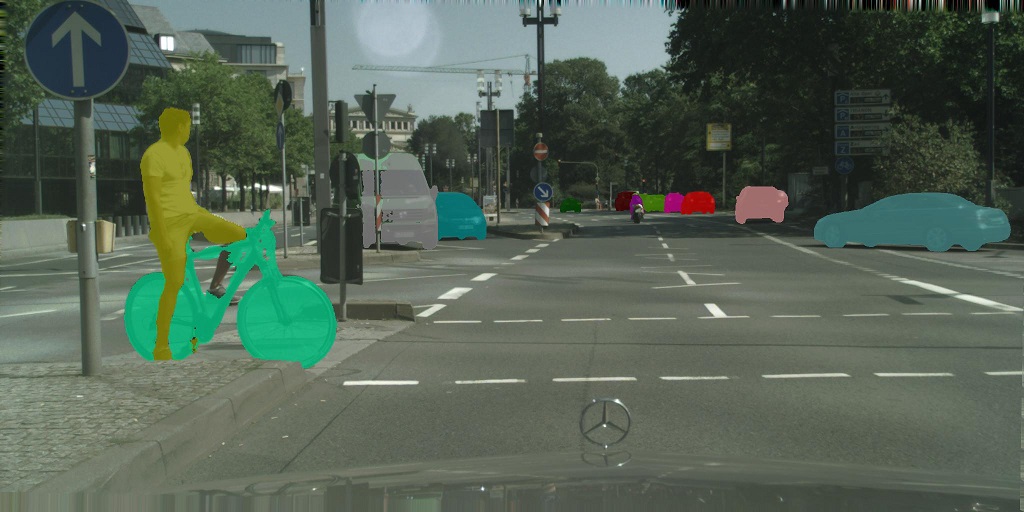}
		\end{minipage}
		\begin{minipage}[t]{0.23\textwidth}
			\centering
			\includegraphics[width=2.8cm]{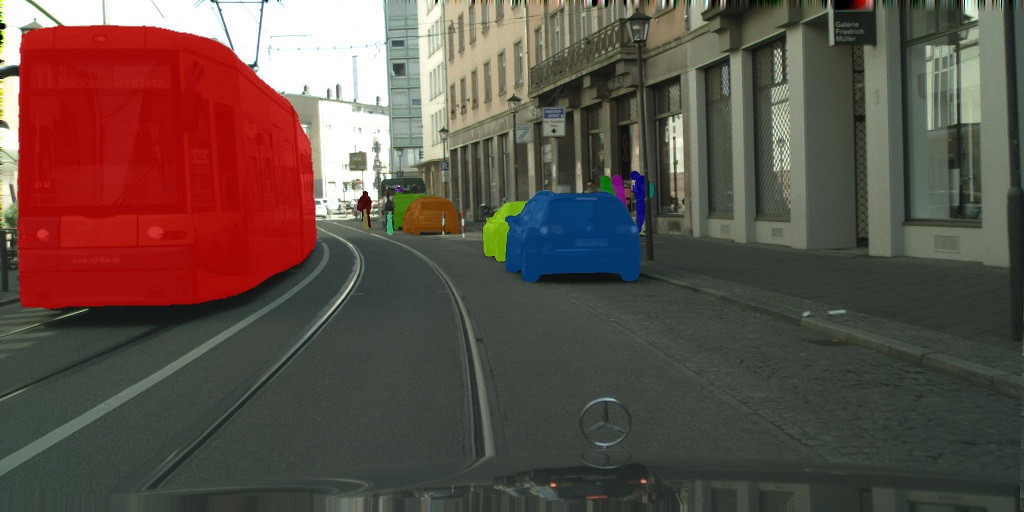}
		\end{minipage}
		\begin{minipage}[t]{0.23\textwidth}
			\centering
			\includegraphics[width=2.8cm]{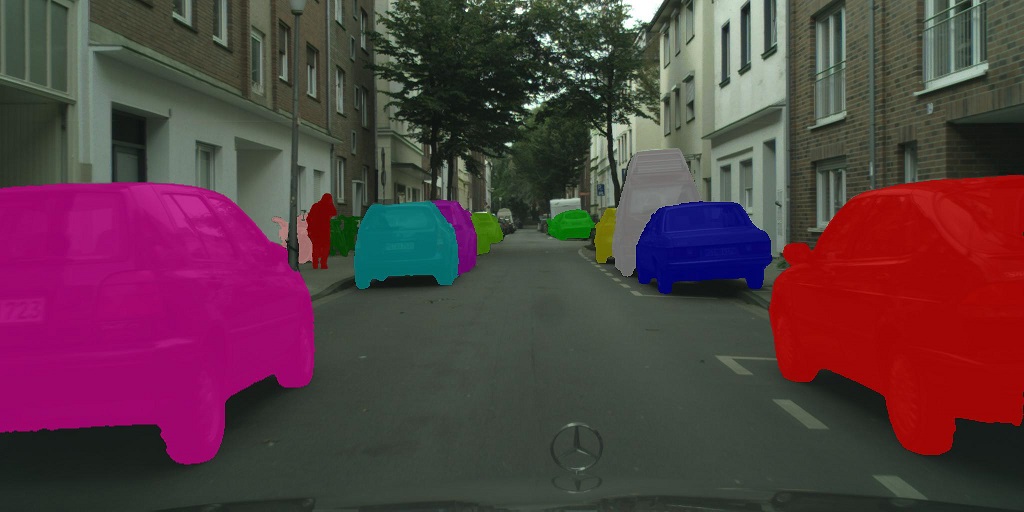}
		\end{minipage}
		\begin{minipage}[t]{0.23\textwidth}
			\centering
			\includegraphics[width=2.8cm]{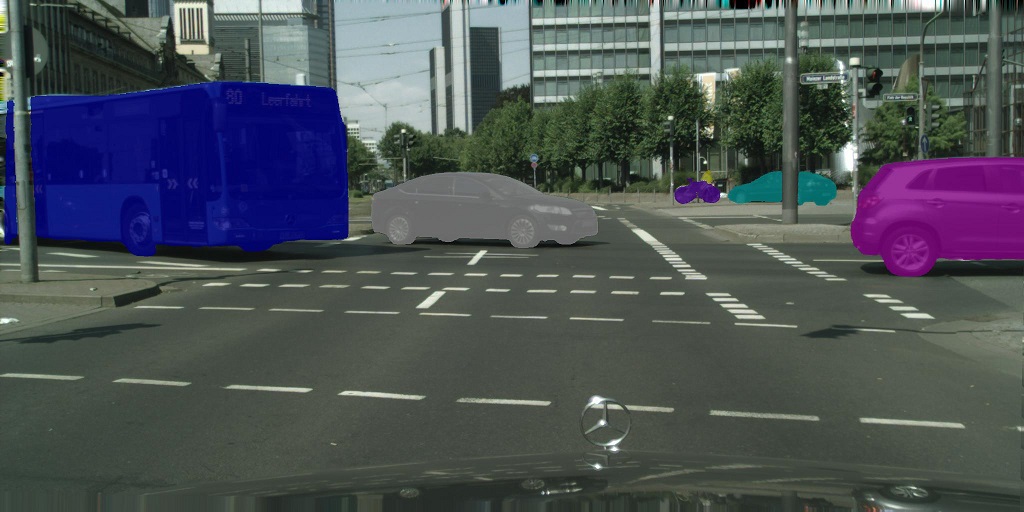}
		\end{minipage}
		\begin{minipage}[t]{0.23\textwidth}
			\centering
			\includegraphics[width=2.8cm]{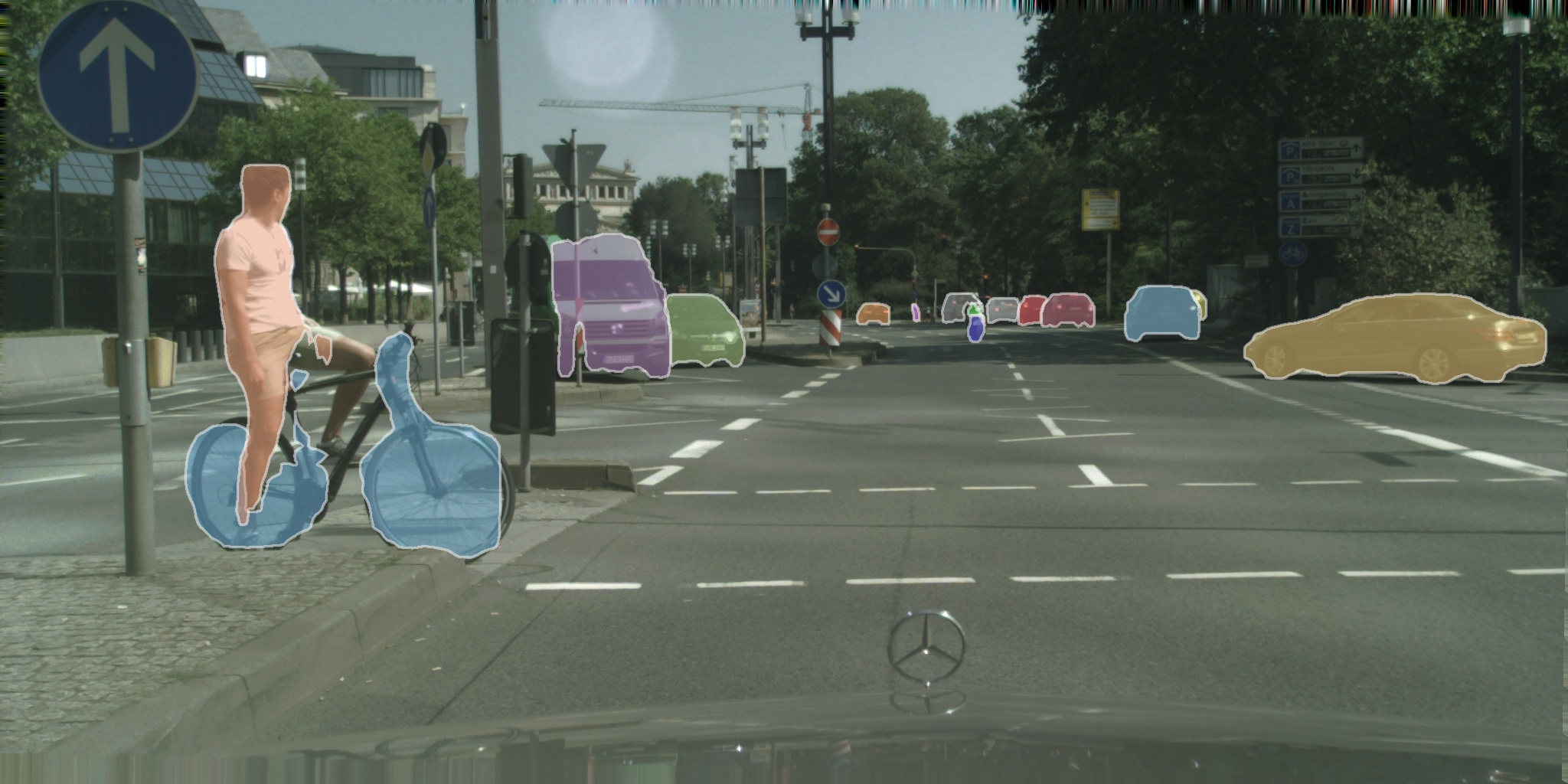}
		\end{minipage}
		\begin{minipage}[t]{0.23\textwidth}
			\centering
			\includegraphics[width=2.8cm]{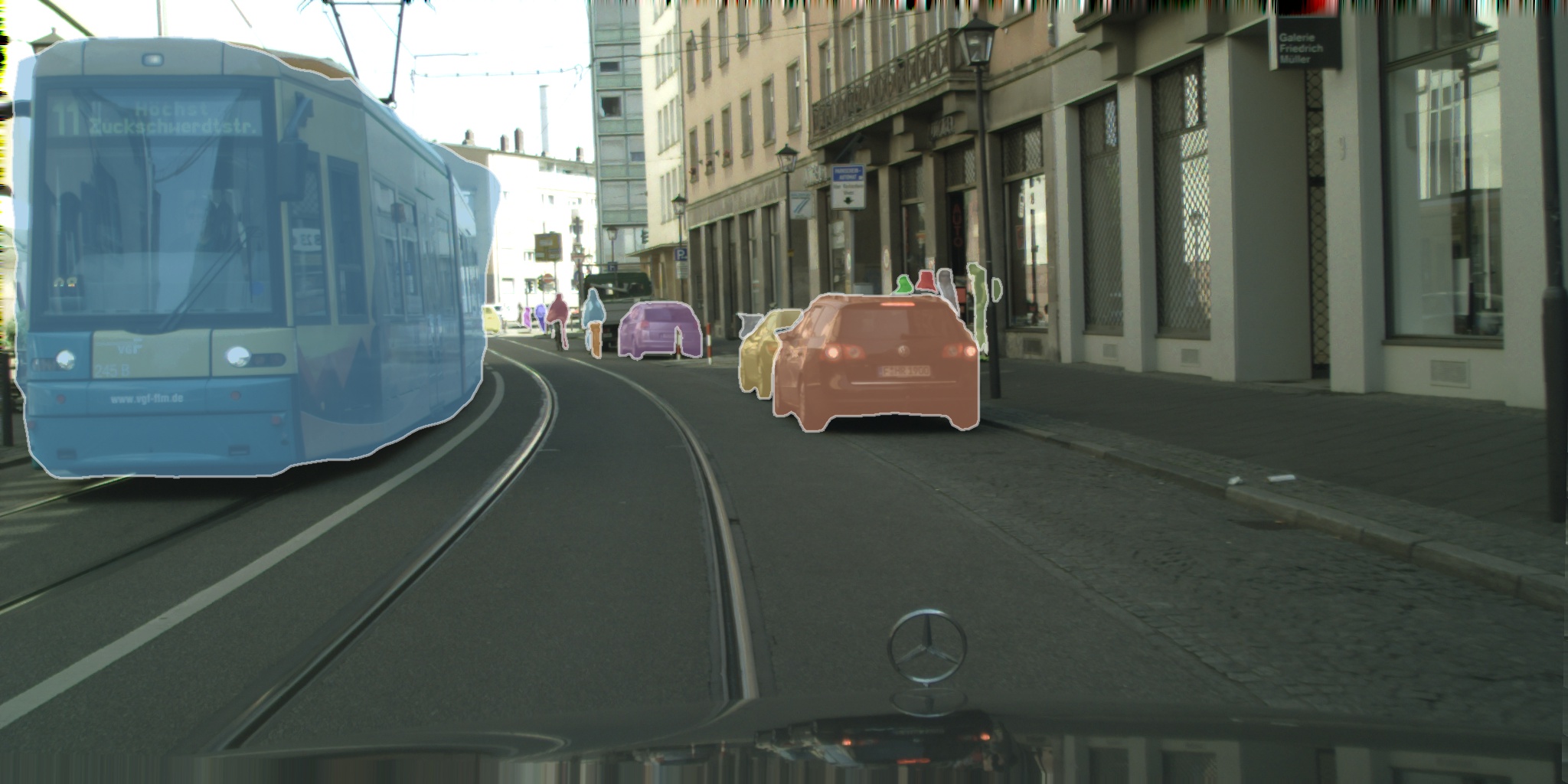}
		\end{minipage}
		\begin{minipage}[t]{0.23\textwidth}
			\centering
			\includegraphics[width=2.8cm]{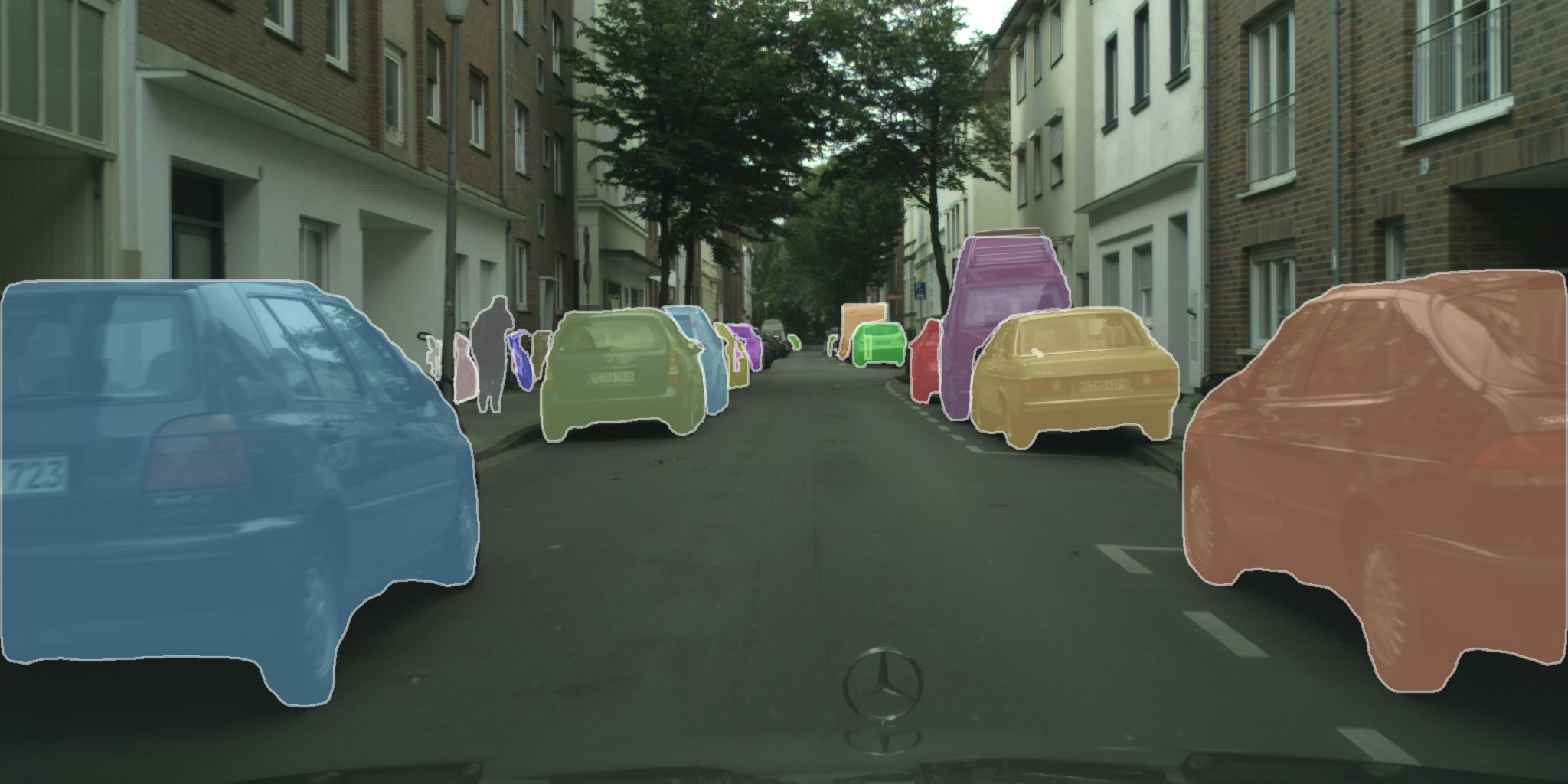}
		\end{minipage}
		\begin{minipage}[t]{0.23\textwidth}
			\centering
			\includegraphics[width=2.8cm]{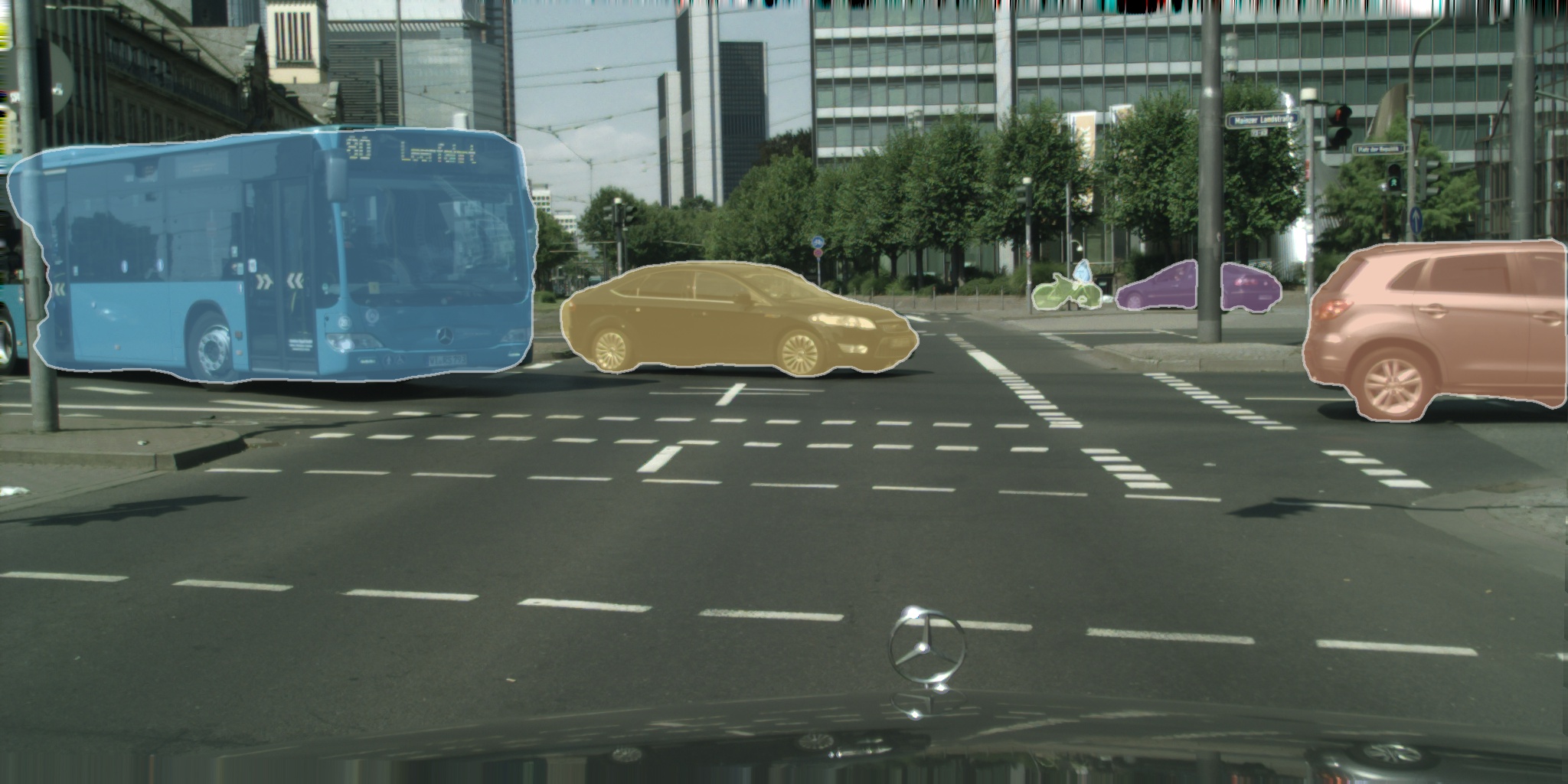}
		\end{minipage}
		
		\caption{Results compared with Mask RCNN. The first row are our results and the second row are results from Mask RCNN. As shown in the figure, we generate more fine-grained masks.}
		\label{fig: compare with Mask RCNN}
	\end{figure}

	Utilizing the implementation of Mask RCNN in Detectron\footnote{https://github.com/facebookresearch/Detectron}, we generate the instance masks and compare them with our results. As shown in Fig. \ref{fig: compare with Mask RCNN}, our results are finer grained. It can be expected that results will be better if we substitute the mask from Mask RCNN with ours when both approaches have prediction of a certain instance. 
	
	\setlength{\tabcolsep}{4pt}
	\begin{table}[t]
		\newcommand{\tabincell}[2]{\begin{tabular}{@{}#1@{}}#2\end{tabular}}
		\begin{center}
			\caption{Instance segmentation performance on \textit{test} set of Cityscapes, all results listed are trained with only Cityscapes dataset}
			\label{table:result on test}
			\resizebox{\textwidth}{15mm}{
				\begin{tabular}{c|c|c|c|c|c|c|c|c|c|c}
					\hline%\noalign{\smallskip}
					Methods & person & rider & car & trunk & bus & train & mcycle & bicycle & AP 50\% & AP \\
					%\noalign{\smallskip}
					\hline
					%\noalign{\smallskip}
					InstanceCut\cite{kirillov2017instancecut}& 10.0 & 8.0  & 23.7 & 14.0 & 19.5 & 15.2 & 9.3  & 4.7  & 27.9 & 13.0\\
					SAIS\cite{hayder2016shape}               & 14.6 & 12.9 & 35.7 & 16.0 & 23.2 & 19.0 & 10.3 & 7.8  & 36.7& 17.4 \\
					DWT\cite{bai2017deep}                    & 15.5 & 14.1 & 31.5 & 22.5 & 27.0 & 22.9 & 13.9 & 8.0  & 35.3& 19.4 \\
					DIN\cite{arnab2017pixelwise}             & 16.5 & 16.7 & 25.7 & 20.6 & 30.0 & 23.4 & 17.1 & 10.1 & 38.8& 20.0 \\
					SGN\cite{liu2017sgn}                     & 21.8 & 20.1 & 39.4 & \textbf{24.8} & 33.2 & \textbf{30.8} & 17.7 & 12.4 & 44.9 & 25.0\\
					Mask RCNN\cite{he2017mask}               & 30.5 & 23.7 & \textbf{46.9} & 22.8 & 32.2 & 18.6 & \textbf{19.1} & \textbf{16.0} & \textbf{49.9} & 26.2\\
					\hline
					Ours                                     & \textbf{31.5} & \textbf{25.2} & 42.3 & 21.8 & \textbf{37.2} & 28.9 & 18.8 & 12.8 & 45.6& \textbf{27.3} \\
					\hline
			\end{tabular}}
		\end{center}
	\end{table}
	\setlength{\tabcolsep}{1.4pt}
	
	\begin{figure}[t]
		\setlength{\belowcaptionskip}{-5pt}
		
		\centering
		\begin{minipage}[t]{0.23\textwidth}
			\centering
			\includegraphics[width=2.8cm]{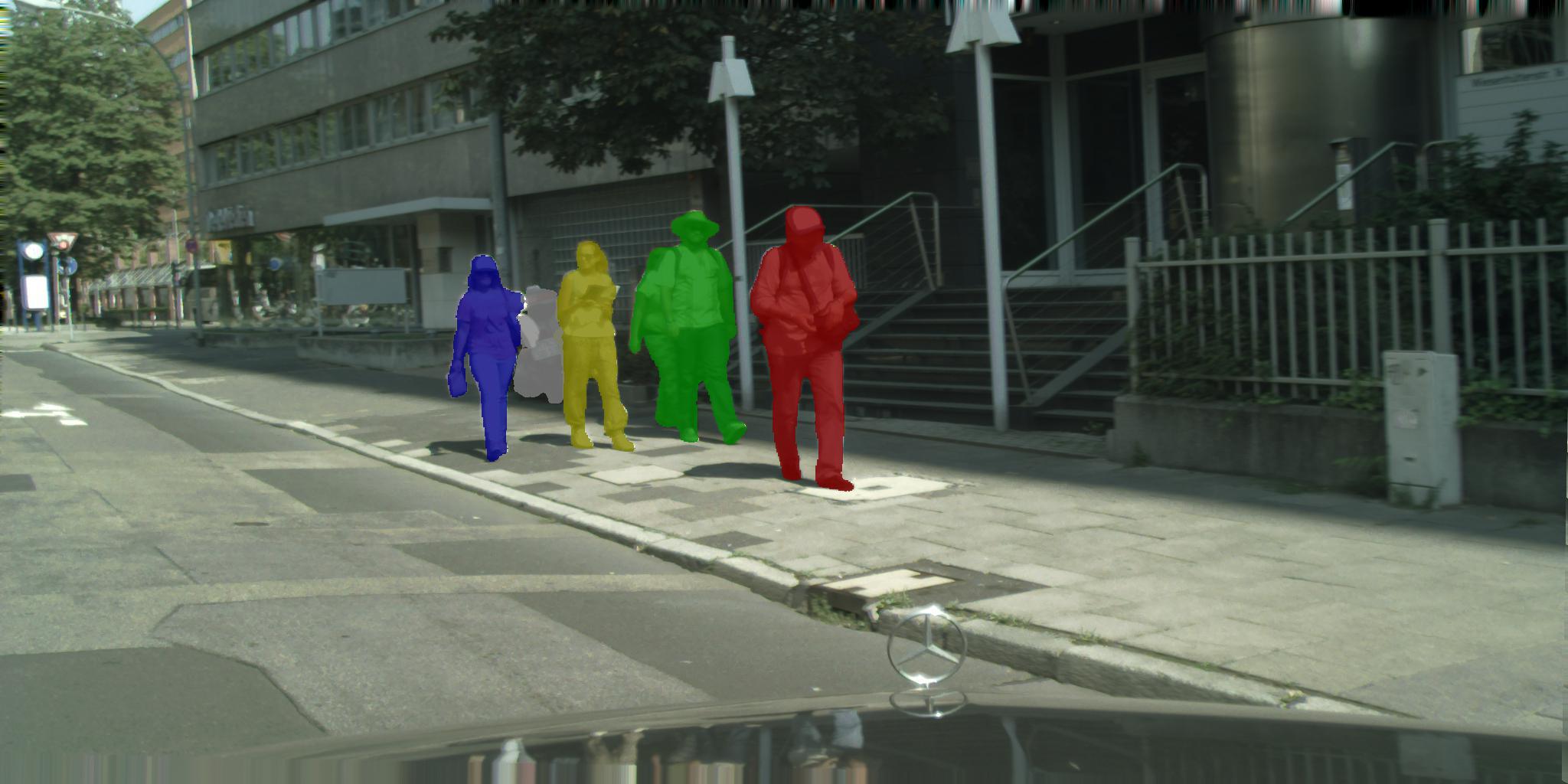}
		\end{minipage}
		\begin{minipage}[t]{0.23\textwidth}
			\centering
			\includegraphics[width=2.8cm]{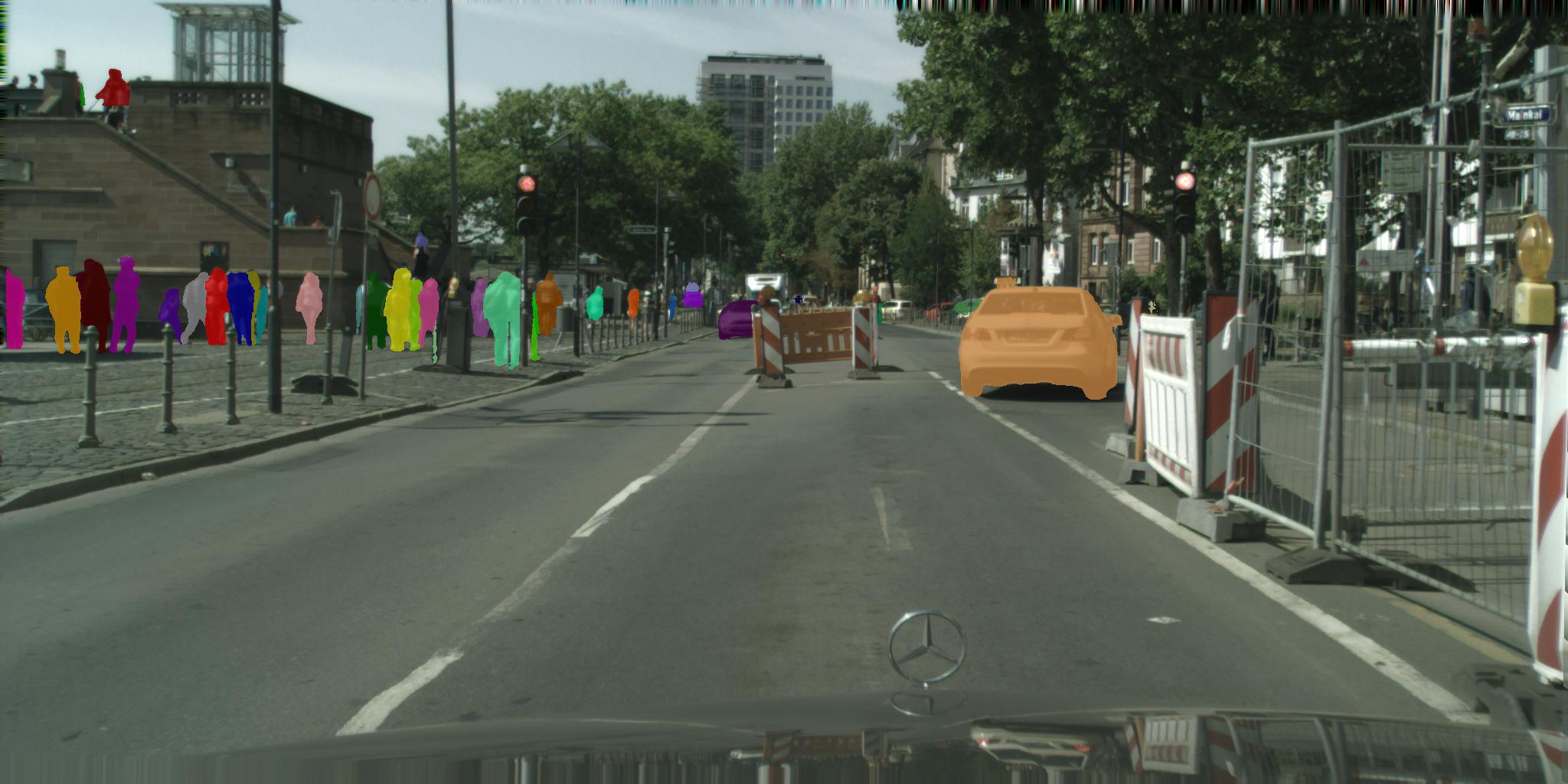}
		\end{minipage}
		\begin{minipage}[t]{0.23\textwidth}
			\centering
			\includegraphics[width=2.8cm]{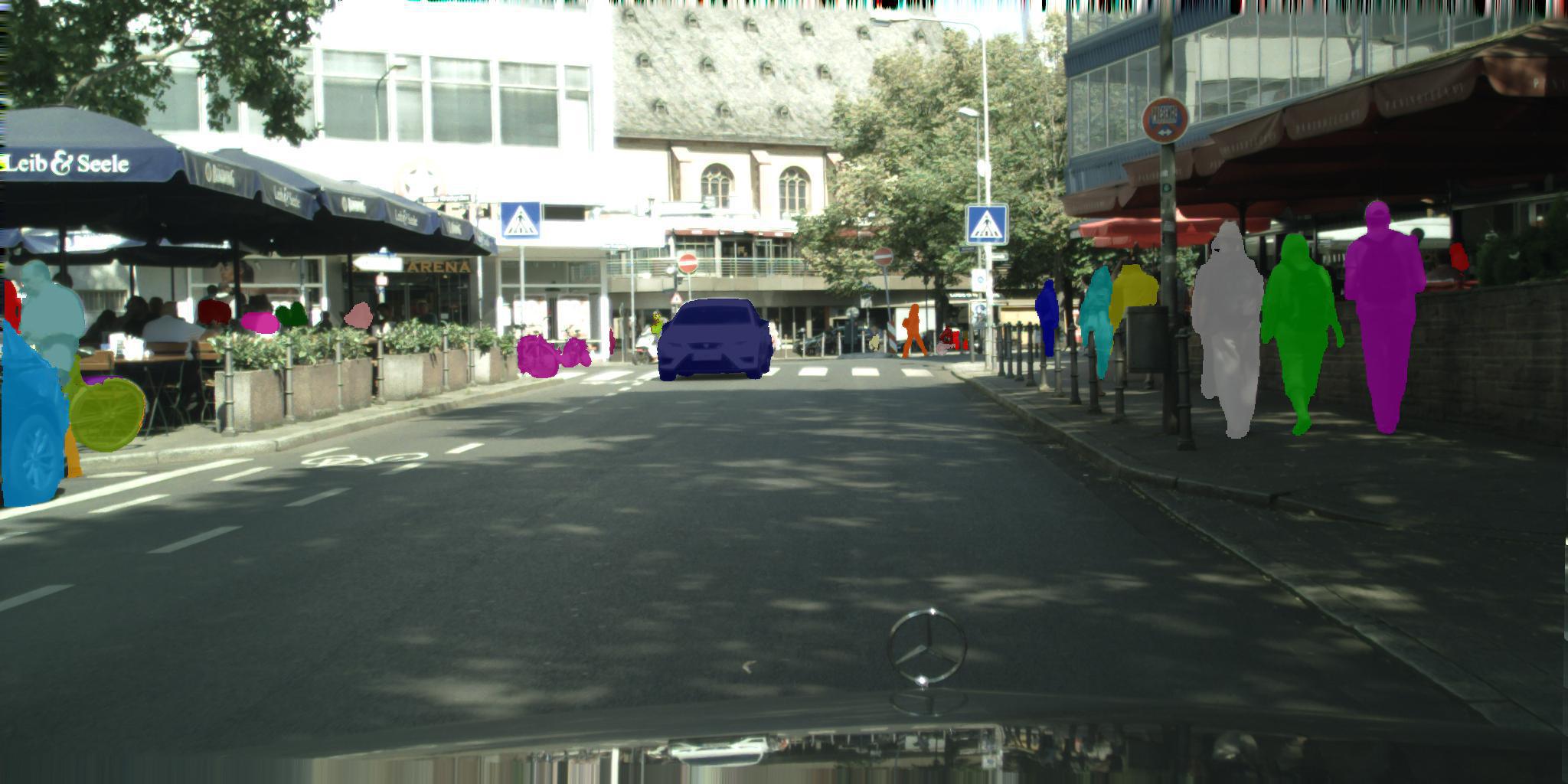}
		\end{minipage}
		\begin{minipage}[t]{0.23\textwidth}
			\centering
			\includegraphics[width=2.8cm]{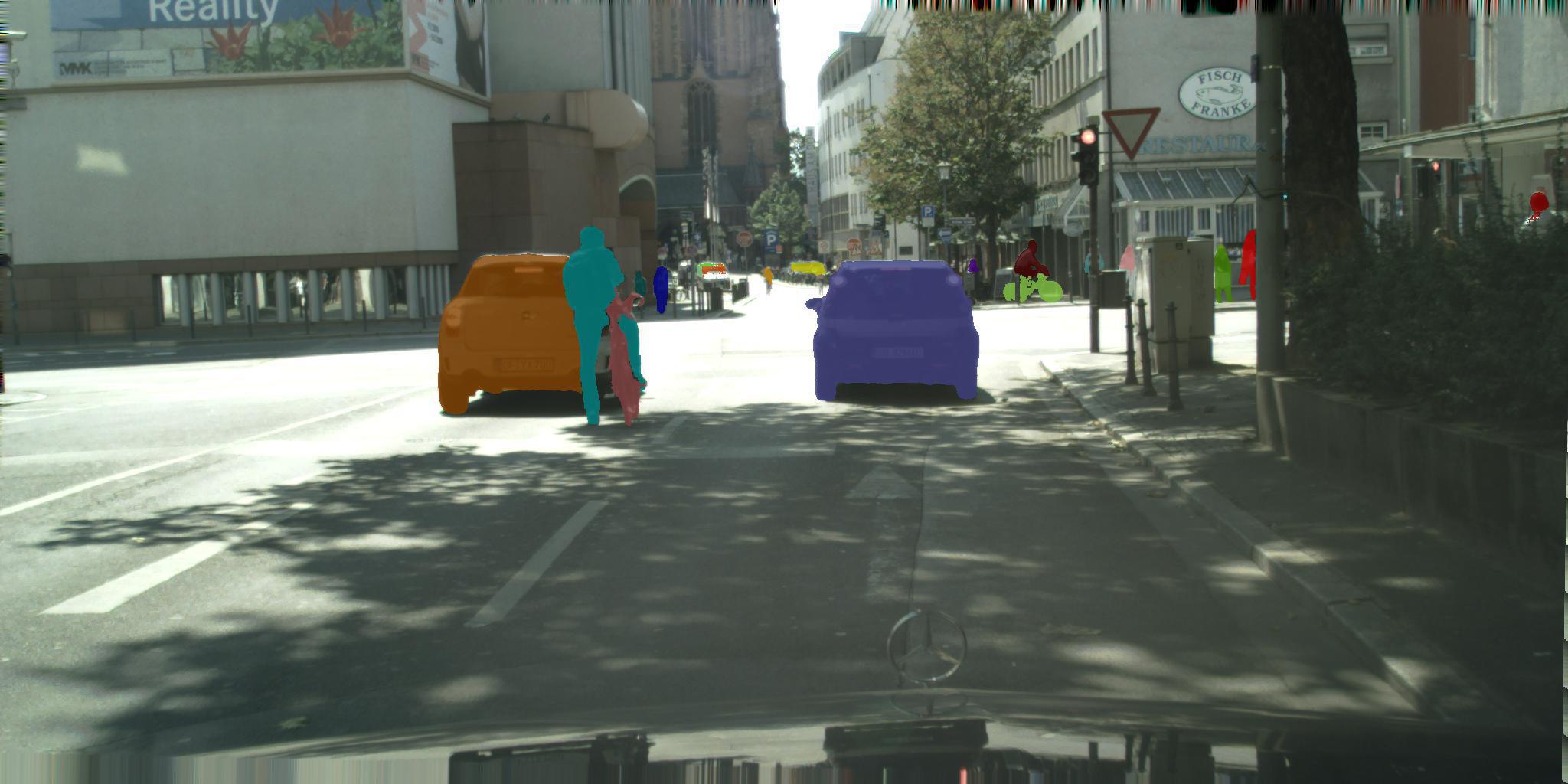}
		\end{minipage}
		
		\caption{Examples of faliure case}
		\label{fig: faliure case}
	\end{figure}

	\subsection{Detailed Results}
	\label{subsec: results}
	We report the ablation studies with \textit{val} set and discuss in detail.

	\textbf{Baseline: }we take the algorithm we describe before Sec. \ref{subsec:excluding backgrounds} as the baseline, for excluding backgrounds helps to significantly speedup the graph merge algorithm and hardly affects the final results. We get 18.9\% AP as our baseline, and we will introduce the results for strategies applied to graph merge algorithm. 
	
	\setlength{\tabcolsep}{4pt}
	\begin{table}[t]
		\newcommand{\tabincell}[2]{\begin{tabular}{@{}#1@{}}#2\end{tabular}}
		\begin{center}
			\caption{\textbf{Graph Merge Strategy:} we test for our graph merge strategies for our algorithm including \textbf{PAR}: Pixel Affinity Refinement, \textbf{RR}: Resizing ROIs, \textbf{FLM}: Forcing Local Merge and \textbf{SCP}: Semantic Class Partition. Note that 2 and 4 in FLM represent the merge window size, default as 1.}
			
			\label{table:graph merge strategy}
			\begin{tabular}{ccccc}
				\noalign{\smallskip}
				\hline
				\noalign{\smallskip}
				PAR       & RR        & FLM       &SCP        &AP \\
				\noalign{\smallskip}
				\hline
				\noalign{\smallskip}
				&           &           &           &18.9 \\
				\checkmark&           &           &           &22.8 \\
				\checkmark& \checkmark&           &           &28.7 \\
				\noalign{\smallskip}
				\hline
				\noalign{\smallskip}
				\checkmark& \checkmark&2          &           &29.2 \\
				\checkmark& \checkmark&4          &           &27.5 \\
				\noalign{\smallskip}
				\hline
				\noalign{\smallskip}
				\checkmark& \checkmark&2          &\checkmark &30.7 \\
			\end{tabular}
		\end{center}
	\end{table}
	\setlength{\tabcolsep}{1.4pt}
	
	We show the experimental results for graph merge strategies in Table. \ref{table:graph merge strategy}.
	For pixel affinity refinement, we add semantic information to refine the probability and get a 22.8\% AP result. As shown in the table, it provides 3.9 points AP improvement. Then we resize the ROIs with a fix size 513, and we get a raise of 5.9 points AP, which significantly improve the result.
	Merge window size influences the result a lot. We have a 0.5 point improvement utilizing window size 2 and a 1.2 point drop with window size 4. As we can see, utilizing 2 as the window size not only reduce the complexity of graph merge, but also get improvement on performance, but utilizing 4 will cause a lost on detailed information and perform below expectation. Therefore, we utilize 2 in the following experiments.
	As mentioned in Sec. \ref{subsec:excluding backgrounds}, we finally divided semantic classes into 3 subclasses for semantic class partition:\{\textit{person, rider}\}, \{\textit{car, trunk, bus, train}\} and \{\textit{motorcycle, bicycle}\}, finding feasible areas separately. Such separation reduces the influence across subclasses and makes the ROI resize more effective.
	We get a 1.5 improvement by applying this technique from 29.0\% to 30.5\%, as shown in the table.
	It is noted that utilizing larger image can make results better, but it also increases the processing time.
	
	Besides the strategies we utilize in the graph merge, we also test our model for different inference strategies referring to \cite{chen2017rethinking}. 
	Output stride is always important for the segmentation-like task.
	Small output stride usually means more detailed information but more inference time cost and smaller batch size in training.
	We test our models firstly trained on output stride 16, then we finetuned models on output stride 8 as in \cite{chen2017rethinking}. It shows in Table. \ref{table:alternative inference strategy} that both semantic and instance model finetuned with output stride 8 improve the result by 0.5 point individually. When combined together, we achieve 32.1\% AP with 1.4 point improvement compared with output stride 16.
	
	We apply horizontal flips and semantic class refinement as alternative inference strategies.
	Horizontal flips for semantic inference brings 0.7 point increase in AP, and for instance inference flip, 0.5 point improvement is observed. We then achieve 33.5\% AP combining these two flips.
	
	\setlength{\tabcolsep}{4pt}
	\begin{table}[t]
		\newcommand{\tabincell}[2]{\begin{tabular}{@{}#1@{}}#2\end{tabular}}
		
		\setlength{\abovecaptionskip}{-5pt}
		\begin{center}
			\caption{\textbf{Additional inference strategies: } We test for additional inference strategies for our algorithm including \textbf{Semantic OS}: output stride for semantic branch, \textbf{Instance OS}: output stride for instance branch \textbf{SHF}: Semantic horizontal flip inference,\textbf{IHF}: Instance horizontal flip inference and \textbf{SCR}: Semantic Class Refinement. We also list several results from other methods for comparison.}
			\label{table:alternative inference strategy}
			\begin{tabular}{c|cccccc}
				\hline
				Methods&Semantic OS & Instance OS & SHF       & IHF       & SCR       & AP \\
				\hline
				DWT\cite{bai2017deep}&&  &          &           &           & 21.2 \\
				SGN\cite{liu2017sgn}&&  &          &           &           & 29.2 \\
				Mask RCNN\cite{he2017mask}&&  &          &           &           & 31.5 \\
				
				\hline
				%\noalign{\smallskip}
				%\hline
				%Ours&&  &          &           &           &  \\
				
				\multirow {8}{*}{Ours}&16& 16 &          &           &           & 30.7 \\
				&8& 16 &           &           &           & 31.2 \\
				&16& 8 &           &           &           & 31.2 \\
				&8& 8 &          &           &           & 32.1 \\
				
				\cline{2-7}
				&8& 8 &\checkmark&           &           & 32.8 \\
				&8& 8 &          & \checkmark&           & 32.6 \\
				&8& 8 &\checkmark& \checkmark&           & 33.5 \\

				&8& 8 &\checkmark& \checkmark& \checkmark& 34.1 \\
				\hline
			\end{tabular}
		\end{center}
		
	\end{table}
	\setlength{\tabcolsep}{1.4pt}
	
	Through observations on the \textit{val} set, we find that instances in \textit{bicycle} and \textit{motorcycle} often fail to be connected when they are fragmentated.
	To improve such situation, we map the pixel affinities between these two classes with Equ.
	\ref{equ:mapping function} at the last distance $d=64$.
	As shown in Table \ref{table:alternative inference strategy}, semantic class refinement get 0.6 point improvement, and get our best result 34.1\% AP on the \textit{val} set.
	
	\subsection{Discussions}
	
	In our current implementation, the maximum distance of the instance branch output is 64. It means that the graph merge algorithm is not able to merge two non-adjacent parts if the distance is greater than 64. Adding more output channels can hardly help the overall performance.
	Moreover, using other network structures, which could achieve better results on the semantic segmentation, may further improve the performance of the proposed graph merge algorithm.	
	Some existing methods, such as \cite{levinkov2017joint}, could solve the graph merge problem but \cite{levinkov2017joint} is much slower than the proposed method. The current graph merge step is implemented on CPU and we believe there is big potential to use multi-core CPU system for acceleration.
	Some examples of failure case are shown in Fig. \ref{fig: faliure case}. The proposed method may miss some small objects or merge different instances together by mistake.
	\section{Conclusions}
	\label{sec:conclusion}
	In this paper, we introduce a proposal-free instance segmentation scheme via affinity derivation and graph merge.
	We generate semantic segmentation results and pixel affinities from two separate networks with a similar structure. Taking these information as input, we regard pixels as vertexes and pixel affinity information as edges to build a graph.
	The proposed graph merge algorithm is then used to cluster the pixels into instances.
	Our method outperforms Mask RCNN on Cityscapes dataset by 1.1 point AP improvement using only Cityscapes training data. It shows that proposal-free method can achieve state-of-the-art performance.
	We notice that the performance of semantic segmentation keep improvement with new methods, which can easily lead to performance improvement for instance segmentation via our method. The proposed graph merge algorithm is simple. We believe that more advanced algorithms can lead to even better performance. Improvements along these directions are left for further work.
	
	\begin{acknowledgement}
		
		Yiding Liu, Wengang Zhou and Houqiang Li's work was supported in part by 973 Program under Contract 2015CB351803, Natural Science Foundation of China (NSFC) under Contract 61390514 and Contract 61632019.
		
	\end{acknowledgement}
	\clearpage
	
	\bibliographystyle{splncs04}
	\bibliography{segmentation}
\end{document}